\documentclass[11pt]{article}

\usepackage[preprint]{acl}

\usepackage{times}
\usepackage{latexsym}

\usepackage[T1]{fontenc}

\usepackage[utf8]{inputenc}

\usepackage{microtype}

\usepackage{inconsolata}

\usepackage{graphicx}
\usepackage{color, colortbl, xcolor}
\usepackage{url}
\usepackage{amsmath}
\usepackage{amssymb}
\usepackage{subcaption}
\usepackage{textcomp}
\usepackage{soul}
\usepackage{multirow}
\usepackage{enumitem}
\usepackage{mathtools}
\usepackage{siunitx}
\usepackage{epigraph}
\usepackage{caption}
\usepackage{subcaption}
\usepackage[table]{xcolor}
\usepackage{colortbl}
\usepackage[table]{xcolor}
\usepackage{pgf}
\usepackage{booktabs} 
\usepackage{longtable,booktabs}
\usepackage{microtype}
\usepackage{float}

\usepackage{arydshln}
\definecolor{linkColor}{RGB}{6,125,233}
\definecolor{green}{rgb}{0.0, 0.65, 0.31}
\definecolor{bleudefrance}{rgb}{0.19, 0.55, 0.91}
\definecolor{ceruleanblue}{rgb}{0.16, 0.32, 0.75}
\definecolor{grey}{HTML}{969696}
\definecolor{violet}{HTML}{756bb1}
\definecolor{dgrey}{HTML}{01665e}
\definecolor{lgrey}{HTML}{5ab4ac}
\definecolor{dgreen}{HTML}{005a32}
\definecolor{purple}{HTML}{ae017e}

\definecolor{editCol}{HTML}{000000}
\definecolor{maskCol}{HTML}{c51b7d}
\definecolor{lrColor}{HTML}{8856a7}
\definecolor{trColor}{HTML}{d01c8b}
\definecolor{ctColor}{HTML}{4dac26}
\definecolor{brickred}{HTML}{f03b20}

\definecolor{improveCol}{HTML}{4dac26}
\definecolor{worsenCol}{HTML}{d01c8b}

\definecolor{DarkBlue}{HTML}{00008B}
\definecolor{mscolor}{HTML}{01665e}
\definecolor{nmscolor}{HTML}{bf812d}
\definecolor{lgreen}{HTML}{ccece6}
\definecolor{dolive}{HTML}{308014}

\colorlet{tablerowcolor4}{gray!50} 

\newcommand*{\textlabel}[2]{%
  \edef\@currentlabel{#1}
  \phantomsection
  #1\label{#2}
}

\colorlet{tableheadcolor}{gray!25} 
\colorlet{tablerowcolor}{gray!10} 
\colorlet{tablerowcolor2}{gray!45} 
\colorlet{tablerowcolor3}{gray!10} 
\newcommand{\rowcollight}{\rowcolor{tablerowcolor3}} %

\newcolumntype{a}{>{\columncolor{tablerowcolor}}r}
\definecolor{aicolor}{HTML}{018571}
\definecolor{occolor}{HTML}{ff7799}
\definecolor{negcolor}{HTML}{CD2990}
\definecolor{poscolor}{HTML}{018571}

\definecolor{aicolor}{HTML}{fc8d62}
\definecolor{occolor}{HTML}{253494}

\newcommand{\gradcell}[1]{%
  \begingroup
  \pgfmathsetmacro{\val}{#1}%
  \def\cellshade{}%

  \ifdim\val pt>0pt
    \pgfmathtruncatemacro{\shade}{%
      min(90,80*pow(min(abs(\val),35)/35,0.7))%
    }%
    \xdef\cellshade{\noexpand\cellcolor{poscolor!\shade!white}}%
  \else
    \ifdim\val pt<0pt
      \pgfmathtruncatemacro{\shade}{%
        min(90,80*pow(min(abs(\val),35)/35,0.7))%
      }%
      \xdef\cellshade{\noexpand\cellcolor{negcolor!\shade!white}}%
    \fi
  \fi

  \cellshade #1%
  \endgroup
}

\newcommand{\hlpos}[1]{%
  {\setlength{\fboxsep}{0pt}\colorbox{poscolor!20}{#1}}%
}
\newcommand{\hlneg}[1]{%
  {\setlength{\fboxsep}{0pt}\colorbox{negcolor!20}{#1}}%
}

\newif{\ifhidecomments}
  \hidecommentsfalse 
\ifhidecomments
    \newcommand{\soorya}[1]{}
    \newcommand{\koustuv}[1]{}
\else
    \newcommand{\soorya}[1]{\textbf{\small\sffamily{\textcolor{DarkBlue}{[#1 -- Soorya]}}}}
    \newcommand{\koustuv}[1]{\textbf{\small\sffamily{\textcolor{purple}{[#1 -- Koustuv]}}}}
  \fi

\newcommand{\arc}{\texttt{ARC-Easy}}
\newcommand{\gsm}{\texttt{GSM8K}}
\newcommand{\mml}{\texttt{MMLU}}

\newcommand{\Num}[1]{$\mathtt{#1}$}

\colorlet{tableheadcolor}{gray!25} 

\definecolor{neutralCol}{HTML}{dd1c77}
\definecolor{neutralGreen}{HTML}{31a354}
\definecolor{NewBlue}{HTML}{1879ba}
\definecolor{bleudefrance}{rgb}{0.19, 0.55, 0.91}  
\definecolor{AfTrColor}{HTML}{0868ac}  
\definecolor{BfTrColor}{HTML}{a8ddb5}  

\definecolor{AfCtColor}{HTML}{b10026}  
\definecolor{BfCtColor}{HTML}{fd8d3c}

\graphicspath{ {figures/} }

\newcommand{\para}[1]{\vspace{0.2em}\noindent\textbf{\textit{#1}~}}

\title{Prompt PerturbAItions: Mechanisms of LLM Response Inconsistencies}

\author{
 \textbf{Soorya Ram Shimgekar\textsuperscript{$\spadesuit$}},~
 \textbf{Agam Goyal\textsuperscript{$\spadesuit$}},~
 \textbf{Amruta Parulekar\textsuperscript{$\spadesuit$}},~
 \textbf{Joshua Chen\textsuperscript{$\spadesuit$}},~
 \textbf{Yian Wang\textsuperscript{$\spadesuit$}},\\
 \textbf{Navin Kumar\textsuperscript{$\diamondsuit$}},~
\textbf{Hari Sundaram\textsuperscript{$\spadesuit$}},~
\textbf{Eshwar Chandrasekharan\textsuperscript{$\spadesuit$}},~
\textbf{Koustuv Saha\textsuperscript{$\spadesuit$}},
\\\\
\textsuperscript{$\diamondsuit$}Nimblemind\\
 \textsuperscript{$\spadesuit$}Siebel School of Computing and Data Science\\  University of Illinois Urbana-Champaign, Urbana, IL, USA
\\
\{sooryas2, agamg2, amp20, joshua86, yian3, hs1, eshwar, ksaha2\}@illinois.edu, navin@nimblemind.ai
}

\begin{document}
\maketitle

\begin{abstract}
Large language models (LLMs) are increasingly deployed in conversational settings where user phrasing ranges from polite to casual to toxic, yet less is known about whether such lexical variation in otherwise semantically equivalent prompts can degrade factual reliability.
We study how lexical and tone-based prompt perturbations affect the factual reliability of LLMs. 
Using controlled prompt variations spanning polite, random, and increasingly toxic levels, we evaluate seven LLMs on \arc{}, \gsm{}, and \mml{}. 
We find that lexical perturbations reduce factual accuracy and increase predictive uncertainty in most models. Interestingly, increasing toxicity results in decreasing accuracy. 
To examine whether these answer inconsistencies correspond to internal changes, we conduct attribution-graph analyses of model activations and influences. 
We find that perturbations amplify perturbation-sensitive variant nodes relative to core reasoning nodes. We also observed a monotonic trend with increasing toxicity score and increasing activation and influence in Variant nodes.
These findings position prompt wording and tone as a critical dimension of LLM reliability and provide behavioral and mechanistic evidence that surface-level lexical variation can alter factual outputs and internal computation. 



\end{abstract}

\section{Introduction}
\label{sec:introduction}



\begin{quote}
\vspace{-0.5em}
\small
``Saying ``please'' or ``thank you'' to AI chatbots can apparently cost tens of millions of dollars. But some fear the cost of not being polite could be higher.''---New York Times~\cite{deb2025saying}
\vspace{-0.5em}
\end{quote}

\noindent As AI systems become increasingly embedded in everyday workflows, users interact with LLMs in various conversational settings, ranging from polite and carefully structured prompts to emotionally charged, adversarial, or toxic language~\cite{gehman2020realtoxicityprompts, wei2023jailbroken}. 
Recent public discourse has drawn attention to the possible practical implications of users' conversational variations, including computational cost, with OpenAI CEO Sam Altman claiming that polite interactions such as please'' and thank you'' cost the company millions of dollars in compute~\cite{futurism2025please}.
At the same time, growing evidence suggests that seemingly minor prompt variations can substantially alter model behavior and downstream performance~\cite{zhao2021calibrate, lu2022fantastically, perez2023discovering, dobariya2025mind, mizrahi2024state,sclar2024quantifying,yin2024should}. 
This raises important questions regarding the robustness and reliability of factual reasoning under variations in prompt wording and tone.

A rich body of work has studied hallucinations in LLMs, typically referring to fluent but factually unsupported or fabricated generations~\cite{ji2023survey, maynez2020faithfulness, dziri2022faithfulness}. 
However, recent research argues that factual failures in structured tasks such as question answering and multiple-choice reasoning may arise from several distinct mechanisms, including reasoning instability, prompt sensitivity, uncertainty, and failures in knowledge elicitation~\cite{jang2026adaptive}.
In particular, semantically equivalent prompts can often produce inconsistent factual answers despite preserving the underlying query intent~\cite{dobariya2025mind, cai2025tone,elazar2021measuring}. 
This distinction matters because prompt-induced factual inconsistency represents a broader reliability challenge that extends beyond conventional notions of hallucination.

Prior work has focused on stylistic prompting strategies such as calibration, role prompting, formatting, and chain-of-thought reasoning~\cite{zhao2021calibrate, lu2022fantastically, perez2023discovering}.
However, the role of everyday lexical and tonal variation in shaping factual reliability, spanning courteous phrasing, irrelevant insertions, and, at the far end, hostile or toxic remarks~\cite{gehman2020realtoxicityprompts,davidson2017automated}, remains underexplored. This gap is especially important because real-world interactions with AI are driven by how users naturally choose to prompt, and these interactions span a wide range of tones, occasionally including adversarial or emotionally aggressive language~\cite{barhoumusers, gehman2020realtoxicityprompts}. 
Therefore, understanding how such variation affects LLM outputs is critical for evaluating the robustness and safety of deployed LLM systems.

Therefore, to empirically examine this phenomenon, we study how lexical perturbations inserted into otherwise semantically equivalent prompts alter the ability of the LLMs 
in factual answering and reasoning behavior.
Rather than treating all incorrect outputs as conventional hallucinations, we focus more precisely on how perturbing prompts---ranging across random, polite, and toxic insertions---could impact factual reliability and answer instability.
Further, recent advances in mechanistic interpretability have emphasized understanding LLMs through their internal representations, often conceptualized as \textit{attribution graphs} or \textit{circuits} responsible for specific behaviors~\cite{elhage2021mathematical, olah2020zoom, geva2021transformer}. 
Building on this line of work, we investigate how lexical perturbation-induced factual instability corresponds to identifiable shifts in internal computation and attribution graph behavior.
Our work is guided by the following research questions (RQs):

    \para{RQ1:} How do lexical and tone-based prompt perturbations impact the factual reliability of LLMs?

    \para{RQ2:} Can such factual degradation be explained through internal representations within LLMs?

Our work examines seven models: GPT-5-nano, Gemini-2.5-Flash, Gemma-2-2B, Qwen2.5-1.5B-Instruct, Qwen2.5-7B-Instruct, LLaMA-3.2-1B, and LLaMA-3-8B-Instruct under prompt perturbations ranging across random, polite, and toxic prompts. 
We evaluate these models and prompt perturbations across three widely used benchmarks: \arc{}~\cite{clark2018think}, \gsm{}~\cite{cobbe2021training}, and \mml{}~\cite{hendrycks2021measuring}.

For RQ1, we conduct lexical perturbations and compare the factual accuracy of the models across the benchmarks.
Regression models examine how various aspects of the prompt associate with the models' accuracy, entropy, and perplexity. 
We find that lexical perturbations degrade factual accuracy in most models, while higher toxicity is associated with increased predictive uncertainty as measured by entropy.
Even random, non-semantic insertions reduce performance, indicating that seemingly innocuous prompt variations can destabilize reasoning behavior, with the effects becoming more pronounced as perturbations grow more toxic.
Degradation varies widely across models, from small drops for GPT-5-nano (-4.21\% under high-toxic prompts) to large drops for Gemma-2-2B (-36.14\%).

For RQ2, 
we trace the circuits of three open-source models (Gemma-2-2B, LLaMA-3.2-1B, and Qwen3-1.7B), specifically, the attribution graphs of activations and influences across layers. 
We find that lexical perturbations amplify perturbation-sensitive pathways, increasing their activation and influence while core reasoning nodes decrease in activation and influence, a shift that grows stronger toward the toxic end of the spectrum. These internal shifts closely align with the observed degradation in factual accuracy and increased uncertainty suggesting that lexical perturbations, particularly toxic ones, redirect computation away from stable semantic reasoning circuits toward context-sensitive representations.

This work makes four contributions: 1) a computational framework for studying lexical and tone-based prompt perturbations in factual reasoning through controlled rewrites and attribution-graph evaluation; 2) a cross-model empirical analysis showing that lexical perturbations, especially toxic ones, can degrade factual reliability and answer consistency; 3) mechanistic insights into perturbation-sensitive attribution subgraphs associated with factual instability; and 4) the release of dataset, prompt perturbation framework, and codebase to support future research on tone-sensitive robustness and mechanistic analysis in LLMs.

\section{Related Work}
\label{sec:related_work}

\para{Factuality and Hallucination in LLM Responses.}
Large language models (LLMs) are known to generate fluent but factually incorrect outputs, commonly referred to as hallucinations~\cite{ji2023survey,maynez2020faithfulness,dziri2022faithfulness}. 
Prior work has characterized hallucinations across tasks such as summarization, question answering, and dialogue, attributing them to factors including spurious correlations in training data, exposure bias, and lack of grounding~\cite{maynez2020faithfulness,shuster2021retrieval}.
Recent work found AI-mediated delusional reinforcement based on validation patterns~\cite{shimgekar2026ai}.
Several approaches have been proposed to detect or mitigate hallucinations, including self-consistency methods~\cite{wang2023selfconsistency}, sampling-based detection~\cite{manakul2023selfcheckgpt}, and retrieval-augmented generation~\cite{lewis2020retrieval}. 
Recent work emphasizes evaluation frameworks that distinguish between factual accuracy, faithfulness, and calibration~\cite{dziri2022faithfulness,kadavath2022language}. 
However, most existing studies assume fixed prompt conditions and do not examine how \textit{linguistic perturbations}, particularly those unrelated to task semantics, influence hallucination behavior. 
Our work extends this line by studying how 
prompt variations systematically affect factual accuracy.

\para{Prompt Sensitivity and Adversarial Inputs.}
A growing body of work reveals that LLMs are highly sensitive to prompt phrasing, even when semantic content is preserved~\cite{zhao2021calibrate,lu2022fantastically,perez2023discovering}. 
Prompt calibration methods show that minor formatting or ordering changes can significantly alter model predictions~\cite{zhao2021calibrate,lu2022fantastically}. 
Similarly, studies on prompt multiplicity reveal that models can produce inconsistent outputs for equivalent queries~\cite{perez2023discovering}. 
Beyond benign variations, adversarial prompting such as universal triggers~\cite{wallace2019universal} and jailbreak attacks~\cite{xie2023selfreminder,wei2023jailbroken} reveal that carefully constructed inputs can induce harmful or incorrect outputs.
Further, prompt sensitivity could decrease with model scale but does not disappear entirely~\cite{zhuo2024prosa}. 
Our work studies lexical perturbations, with toxicity as a controlled, graded dimension, and examines their effect on LLM responses.

\para{Toxicity, Bias, and Safety in LLMs.}
LLMs inherit biases and toxic patterns from their training data, leading to concerns about fairness, safety, and harmful content generation~\cite{bender2021dangers,weidinger2022taxonomy,goel2026rubrix,kim2026pair}. 
Extensive work has focused on detecting and mitigating toxicity using alignment techniques such as reinforcement learning from human feedback (RLHF)~\cite{ouyang2022training}, decoding-time interventions~\cite{liu-etal-2021-dexperts,zhang-wan-2023-mil}, and interpretability-based activation steering and model editing~\cite{uppaal2025model,goyal-etal-2025-breaking}. 
Safety research also examines how models respond to harmful or adversarial inputs, showing that toxicity can interact with alignment mechanisms in complex ways~\cite{zhou2024alignment,xie2023selfreminder}. 
For example, alignment can operate through intermediate representations that detect harmful intent in early layers and refine responses in later layers~\cite{zhou2024alignment}. 
This body of work primarily treats toxicity as an \textit{output} concern (i.e., preventing harmful outputs).
Parallelly, prior work has explored LLMs in identifying and flagging toxic content from a content moderation perspective~\cite{kolla2024llm,kumar2024watch,goyal2025momoe,zhan2025slm}.
Our work complements this research by examining the effect of toxic tokens as \textit{input} to models.

\para{Mechanistic Interpretability and Circuits in Transformers.}
Recent advances in mechanistic interpretability aim to explain LLM behavior by identifying internal computational structures, often referred to as circuits~\cite{olah2020zoom,elhage2021mathematical}. 
These approaches analyze how specific neurons, attention heads, or subspaces contribute to model outputs. 
For example, prior work shows that feed-forward layers act as key-value memories storing factual associations~\cite{geva2021transformer}, while attention heads can implement structured reasoning patterns. 
Probing methods have been widely used to extract interpretable signals from hidden representations, demonstrating that intermediate activations encode linguistic, factual, and safety-related information~\cite{belinkov2017neural}. 
More recent work applies causal interventions and attribution techniques to identify neurons responsible for specific behaviors, such as factual recall or bias~\cite{meng2022locating}, and toxicity~\cite{wang2026causaldetox}. 
Our approach builds on this paradigm by using attribution graphs with sparse transcoders to identify \textit{sub-circuits} associated with perturbation-induced factual unreliability
\section{Data}


We build a controlled multi-domain benchmark spanning factual recall, commonsense reasoning, and mathematical problem solving, using three widely adopted NLP benchmarks: 1) \arc{} (elementary scientific and factual reasoning)~\cite{clark2018think}, 2) \gsm{} (multi-step arithmetic reasoning)~\cite{cobbe2021training}, and 3) \mml{} (language understanding and commonsense reasoning across academic subjects)~\cite{hendrycks2021measuring}.
These capture complementary reasoning capabilities, allowing us to evaluate whether perturbation-induced response inconsistencies generalize across task types.

\para{Building the evaluation dataset.}
We construct filtered subsets from each benchmark to support standardized short-form evaluation across models.
For \mml{} and \arc{}, we retain only open-ended question--answer pairs whose gold answers consist of a single token, while excluding examples containing multi-token answers. 
For \gsm{}, we only retain examples with valid integer answers. 
These filtering steps reduce variability introduced by free-form multi-token generation and enable consistent answer normalization across models.
Following filtering, we sample 300 question--answer pairs from each dataset (900 in total).

\begin{table}[t]
\centering
\small
\renewcommand*{\arraystretch}{0.85}
\resizebox{\columnwidth}{!}{
\begin{tabular}{l l l}
\textbf{Bin} & \textbf{Toxicity Range} & \textbf{Example perturbation words} \\
\toprule

\textit{Baseline} & 0.0--0.05 & -- \\

\rowcollight \textit{Polite} & 0.0--0.05 & please, kindly, thanks, appreciate \\

\textit{Random} & 0.0--0.25 & banana, quantum, purple, giraffe \\

\rowcollight \textit{Low Toxic} & 0.25--0.50 & annoying, silly, weird, lame \\

\textit{Medium Toxic} & 0.50--0.80 & idiot, moron, worthless, awful \\

\rowcollight \textit{High Toxic} & 0.80+ & bitch, fucker, shithead, asshole \\

\end{tabular}
}
\caption{Lexical perturbation bins with  Perspective API toxicity score ranges (0--1).}
\label{tab:toxic_bins}
\vspace{-1em}
\end{table}


\para{Constructing lexical perturbation bins.}
To analyze the effect of toxicity, we construct prompt perturbations using curated lexical bins grouped by toxicity intensity. 
In our main setting, each perturbation consists of a single word appended to the end of the original question, preserving its semantic intent and ground-truth answer. 
Toxicity scores are obtained using the Perspective API~\cite{perspectiveapi2017}, which provides continuous toxicity estimates. 
Based on these scores, we partition perturbation tokens into five bins corresponding to increasing toxicity levels: 1) \textbf{polite} (consisting of courteous or cooperative words), 2) \textbf{random} (tone-wise neutral words), 3) \textbf{low toxic} (mildly negative or dismissive words), 4) \textbf{medium toxic} (direct insults or derogatory words), and 5) \textbf{high toxic} (strongly profane or aggressive words) (see~\autoref{tab:toxic_bins}).
We identify 100 tokens per bin.

\para{Generating perturbed prompts.}
The baseline condition contains no appended token, while perturbed conditions use tokens from the polite, random, low-toxic, medium-toxic, and high-toxic bins.
That is, for each question from our datasets, we construct multiple prompt variants by appending a single perturbation token from a bin to the original query. 
Our rationale for using a single-token perturbation is that it introduces a minimal and conservative change to the prompt. More complex perturbations would make comparisons harder to interpret, as any observed output change would be more difficult to attribute to a specific prompt change.
However, to examine the effect of position, we also insert the perturbation word in the middle of the question or prepend it to the question (\autoref{tab:perturbation_position_toxicity}).
Our perturbations preserve the semantic meaning of the question while altering only its tone or toxicity, enabling controlled analysis of how non-semantic perturbations affect factual reasoning and answer consistency in LLMs. 

\section{RQ1: Prompt Perturbation-Induced Answer Inconsistencies of LLMs}
\label{sec:RQ1}

\begin{table*}[t]
\centering
\footnotesize
\renewcommand*{\arraystretch}{0.9}
\setlength{\tabcolsep}{2pt}
\resizebox{\textwidth}{!}{%
\begin{tabular}{lcccccccccccccccccccccccccccc}
& \multicolumn{4}{c}{\textbf{GPT-5-nano}}
& \multicolumn{4}{c}{\textbf{Gemini-2.5-Flash}}
& \multicolumn{4}{c}{\textbf{Gemma-2-2B}}
& \multicolumn{4}{c}{\textbf{Qwen2.5-1.5B}}
& \multicolumn{4}{c}{\textbf{LLaMA-3.2-1B}}
& \multicolumn{4}{c}{\textbf{LLaMA-3-8B-Instruct}}
& \multicolumn{4}{c}{\textbf{Qwen2.5-7B-Instruct}} \\

\cmidrule(r){2-5}
\cmidrule(lr){6-9}
\cmidrule(lr){10-13}
\cmidrule(lr){14-17}
\cmidrule(lr){18-21}
\cmidrule(lr){22-25}
\cmidrule(l){26-29}

\textbf{Bin}
& \textbf{Acc.} & \textbf{$\Delta$\%} & \textbf{$d$} & \textbf{$t$}
& \textbf{Acc.} & \textbf{$\Delta$\%} & \textbf{$d$} & \textbf{$t$}
& \textbf{Acc.} & \textbf{$\Delta$\%} & \textbf{$d$} & \textbf{$t$}
& \textbf{Acc.} & \textbf{$\Delta$\%} & \textbf{$d$} & \textbf{$t$}
& \textbf{Acc.} & \textbf{$\Delta$\%} & \textbf{$d$} & \textbf{$t$}
& \textbf{Acc.} & \textbf{$\Delta$\%} & \textbf{$d$} & \textbf{$t$}
& \textbf{Acc.} & \textbf{$\Delta$\%} & \textbf{$d$} & \textbf{$t$} \\

\midrule

Baseline
& 0.6472 & \gradcell{0.00} & --- & ---
& 0.5906 & \gradcell{0.00} & --- & ---
& 0.2444 & \gradcell{0.00} & --- & ---
& 0.2600 & \gradcell{0.00} & --- & ---
& 0.1111 & \gradcell{0.00} & --- & ---
& 0.2867 & \gradcell{0.00} & --- & ---
& 0.3389 & \gradcell{0.00} & --- & --- \\

Polite
& 0.6328 & \gradcell{-2.23} & -0.0724 & -2.17*
& 0.5994 & \gradcell{+1.51} & 0.0311 & 0.93
& 0.1989 & \gradcell{-18.64} & -0.1614 & -4.84***
& 0.2483 & \gradcell{-4.49} & -0.0494 & -1.48
& 0.1178 & \gradcell{+6.00} & 0.0271 & 0.81
& 0.2461 & \gradcell{-14.15} & -0.1511 & -4.53***
& 0.3233 & \gradcell{-4.59} & -0.0693 & -2.08* \\

Random
& 0.6028 & \gradcell{-6.87} & -0.1801 & -5.40***
& 0.4589 & \gradcell{-22.30} & -0.3557 & -10.67***
& 0.1572 & \gradcell{-35.68} & -0.2636 & -7.91***
& 0.1950 & \gradcell{-25.00} & -0.1990 & -5.97***
& 0.1000 & \gradcell{-10.00} & -0.0487 & -1.46
& 0.2506 & \gradcell{-12.60} & -0.1160 & -3.48***
& 0.3044 & \gradcell{-10.16} & -0.1114 & -3.34*** \\

Low Toxic
& 0.6294 & \gradcell{-2.75} & -0.0862 & -2.59**
& 0.5956 & \gradcell{+0.85} & 0.0171 & 0.51
& 0.1594 & \gradcell{-34.77} & -0.2496 & -7.49***
& 0.2444 & \gradcell{-5.98} & -0.0527 & -1.58
& 0.1150 & \gradcell{+3.50} & 0.0158 & 0.47
& 0.2778 & \gradcell{-3.10} & -0.0347 & -1.04
& 0.2978 & \gradcell{-12.13} & -0.1419 & -4.26*** \\

Medium Toxic
& 0.6244 & \gradcell{-3.52} & -0.1175 & -3.52***
& 0.5628 & \gradcell{-4.70} & -0.0918 & -2.75**
& 0.1694 & \gradcell{-30.68} & -0.2375 & -7.12***
& 0.2239 & \gradcell{-13.89} & -0.1241 & -3.72***
& 0.1161 & \gradcell{+4.50} & 0.0208 & 0.63
& 0.2628 & \gradcell{-8.33} & -0.0846 & -2.54*
& 0.3239 & \gradcell{-4.43} & -0.0547 & -1.64 \\

High Toxic
& 0.6200 & \gradcell{-4.21} & -0.1273 & -3.82***
& 0.4694 & \gradcell{-20.51} & -0.3473 & -10.42***
& 0.1561 & \gradcell{-36.14} & -0.2736 & -8.21***
& 0.2206 & \gradcell{-15.17} & -0.1328 & -3.98***
& 0.1183 & \gradcell{+6.50} & 0.0319 & 0.96
& 0.2611 & \gradcell{-8.91} & -0.0913 & -2.74**
& 0.3083 & \gradcell{-9.02} & -0.1052 & -3.15** \\

\end{tabular}%
}
\caption{Model-wise accuracy (Acc.) under lexical perturbation conditions relative to the baseline, with $\Delta$\% denoting percentage change in accuracy relative to baseline (\hlpos{teal}: positive; \hlneg{pink}: negative; shading indicates magnitude), Cohen's $d$, and $t$-tests ($*$ $p<0.05$, $**$ $p<0.01$, $***$ $p<0.001$). Values are averaged across datasets; dataset-wise results are in Tables~\ref{tab:model_accuracy_comparison_arc_easy}, \ref{tab:model_accuracy_comparison_gsm8k}, and~\ref{tab:model_accuracy_comparison_mmlu}.\vspace{-12pt}
}
\label{tab:model_accuracy_comparison}

\end{table*}

\begin{table}[t]
\centering
\tiny
\setlength{\tabcolsep}{2pt}
\renewcommand*{\arraystretch}{0.9}
\resizebox{\columnwidth}{!}{
\begin{tabular}{llllllllll}
& \multicolumn{3}{c}{\textbf{\arc{}}}
& \multicolumn{3}{c}{\textbf{\gsm{}}}
& \multicolumn{3}{c}{\textbf{\mml{}}} \\

\cmidrule(r){2-4}
\cmidrule(r){5-7}
\cmidrule(l){8-10}

\textbf{Indep. Var.}
& \textbf{Accuracy} & \textbf{Entropy} & \textbf{Perplexity}
& \textbf{Accuracy} & \textbf{Entropy} & \textbf{Perplexity}
& \textbf{Accuracy} & \textbf{Entropy} & \textbf{Perplexity} \\

\midrule

GPT-5-nano
& 0.018* & --- & ---
& 0.883*** & --- & ---
& 0.283*** & --- & --- \\

Gemini-2.5-Flash
& 0.021* & --- & ---
& 0.636*** & --- & ---
& 0.281*** & --- & --- \\

Gemma-2-2B
& -0.031*** & 0.629*** & -5.349***
& -0.055*** & 0.834*** & -0.470***
& -0.077*** & 0.596*** & -4.840*** \\

Llama-3-8B-Instruct
& 0.044*** & -0.889*** & -5.951***
& 0.039*** & -0.496*** & -1.131***
& 0.015* & -0.708*** & -5.394*** \\

Llama-3.2-1B
& -0.124*** & 0.782*** & -5.300***
& -0.025*** & 1.624*** & -0.244***
& -0.200*** & 0.627*** & -4.792*** \\

Qwen2.5-7B-Instruct
& 0.035*** & -0.803*** & -4.834***
& 0.118*** & -0.942*** & -1.229***
& 0.100*** & -0.772*** & -4.806*** \\

Toxicity Score
& -0.004* & 0.071*** & -0.020*
& -0.004* & 0.013*** & -0.001*
& -0.003* & 0.059*** & -0.002 \\

Question Length
& -0.000* & 0.009* & -0.005*
& -0.028*** & 0.088*** & 0.182***
& -0.015** & 0.074*** & 0.007* \\

Answer Rarity
& -0.134*** & 0.073*** & 0.032*
& -0.045*** & -0.064*** & -0.107***
& -0.020* & 0.027* & 0.067* \\

\hdashline

\rowcollight $R^2$ (marginal)
& 0.094 & 0.563 & 0.460
& 0.591 & 0.772 & 0.143
& 0.125 & 0.482 & 0.428 \\

\rowcollight $R^2$ (conditional)
& 0.505 & 0.686 & 0.513
& 0.638 & 0.821 & 0.208
& 0.468 & 0.591 & 0.484 \\

\end{tabular}}
\caption{Dataset-wise mixed-effects regression coefficients for accuracy, entropy, and perplexity across ARC-Easy, GSM8K, and MMLU 
(* $p$<0.05, ** $p$<0.01, *** $p$<0.001). 
Missing entries indicate metrics unavailable for closed-source models.\vspace{-12pt}}
\label{tab:datasetwise_accuracy_entropy_perplexity_regression}
\end{table}


\subsection{RQ1: Methodology}

For our study, we examine seven models from five model families, ranging across a variety of architectures, training datasets, and number of parameters: GPT-5-Nano, Gemini-2.5-Flash, Gemma-2-2B, Qwen2.5-1.5B-Instruct, Qwen2.5-7B-Instruct, LLaMA-3.2-1B, and LLaMA-3-8B-Instruct. 

\para{Answer Generation.} To ensure standardized evaluation across datasets and models, all models are constrained to generate only one token as output. 
Generation is performed greedily with deterministic decoding ($T$=0), minimizing stochastic variation and enabling comparison across perturbations.
As a robustness check, we also evaluate models under free-form generation (\autoref{tab:model_mean_comparison_free_form}), where the effects of perturbations are smaller and mostly non-significant.
Given a model $f_\theta$ and prompt $p_{i,j}$, the generated prediction is defined as: $\hat{y}_{i,j} = f_\theta(p_{i,j})$, where $\hat{y}_{i,j}$ denotes the model output for question $q_i$ under perturbation token $w_j$. 
Generation is performed across all perturbed prompts, producing a complete set of outputs spanning the bins.

\para{Comparing Factual Reliabilities} 
We compute accuracy for each model under every perturbation and compare it against the baseline setting. 
Accuracy is defined as the proportion of correctly answered questions.
For each model, we then obtain effect size (Cohen's $d$) and paired $t$-tests between the baseline and each perturbed condition, allowing us to measure whether the perturbation led to a significant change in performance. 

\para{Regression Modeling.}
To move beyond just accuracy comparisons and better characterize the factors associated with answer consistencies, we conduct regression modeling.
In particular, we build separate linear mixed-effects regression models for three dependent variables: 1) \textit{accuracy}, capturing task correctness; 2) \textit{entropy}, reflecting uncertainty in the model’s output distribution; and 3) \textit{perplexity}, measuring the confidence of the generated response~\cite{jelinek1977perplexity,shannon1948mathematical}.
The entropy and perplexity could only be obtained for the open-source models (i.e., not for GPT/Gemini).
For independent variables, we use model type, Perspective API-based toxicity score of the perturbed word,
length of question, and rarity of answer (computed using a TF-IDF-based rarity score over the answer vocabulary, to account for distributionally infrequent or uncommon target answers).
Model fit is assessed using marginal and conditional $R^2$.

\subsection{RQ1: Results}

\para{Impact of perturbations on model accuracy.}
\autoref{tab:model_accuracy_comparison} summarizes model accuracy across perturbation bins relative to the baseline. 
In five of seven models (all except Gemini-2.5-Flash and LLaMA-3.2-1B), baseline prompts achieve the highest accuracy, suggesting that even small lexical perturbations can disrupt model behavior. 

We first observe that polite perturbations do not improve accuracy across most models. 
In fact, accuracy drops in GPT-5-nano by -2.23\%, Gemma-2-2B by -18.64\%, LLaMA-3-8B-Instruct by -14.15\%, Qwen2.5-7B-Instruct by -4.59\%, and Qwen2.5-1.5B by -4.49\% relative to the baseline condition. 
The only exceptions are LLaMA-3.2-1B (+6.00\%) and Gemini-2.5-Flash (+1.51\%), neither of which is statistically significant.
These findings suggest that courteous lexical modifiers alone do not consistently improve factual reliability. 

We next observe that random perturbations reduce accuracy across all evaluated models: accuracy drops in GPT-5-nano by -6.87\%, Gemini by -22.30\%, Gemma-2-2B by -35.68\%, Qwen2.5-1.5B by -25.00\%, LLaMA-3-8B-Instruct by -12.60\%, Qwen2.5-7B-Instruct by -10.16\%, and LLaMA-3.2-1B by -10.00\% (not significant) relative to the baseline. 
Interestingly, random perturbations produce larger degradation than low-toxic perturbations in six of seven models (all except Qwen2.5-7B). 
For example, GPT-5-nano decreases more under random perturbations (-6.87\%) than under low-toxic prompts (-2.75\%), with a similar pattern observed for Gemini (-22.30\% vs. +0.85\%). 
This suggests that prompt disruption can affect responses even without explicit toxic content. 

Within toxic perturbations, degradation grows with toxicity in four of seven models. GPT-5-nano exhibits a progressive decline across low-toxic (-2.75\%), medium-toxic (-3.52\%), and high-toxic (-4.21\%) conditions, with similar trends observed for Gemini (+0.85\%, -4.70\%, and -20.51\%), Qwen2.5-1.5B (-5.98\%, -13.89\%, and -15.17\%), and LLaMA-3-8B-Instruct (-3.10\%, -8.33\%, and -8.91\%). Gemma-2-2B degrades heavily at every level (-34.77\%, -30.68\%, and -36.14\%) without a monotonic trend.
These results suggest that toxicity compounds an already existing sensitivity to lexical perturbation, leading to increasingly unstable model predictions. 
These drops are not driven by refusals, which remain below 2.2\% for all models and datasets (\autoref{tab:refusal_rates}).
Finally, we observe substantial differences across models. 
The smallest open-source models (Gemma-2-2B, Qwen2.5-1.5B) show the largest proportional degradation, although the pattern is not strictly tied to scale or openness: Gemini-2.5-Flash drops by -20.51\% under high-toxic prompts, more than LLaMA-3-8B-Instruct and Qwen2.5-7B-Instruct, while LLaMA-3.2-1B does not degrade. 
We also vary where toxic content is placed (appended, inserted mid-input, or prepended; \autoref{tab:perturbation_position_toxicity}).
For most models, accuracy drops grow with toxicity at every position.
Middle insertion is the most consistently harmful position for open-source models (6--20\% drops even at low toxicity), while prepending is least harmful at low toxicity but still degrades accuracy at high toxicity (up to $-20.39\%$).
This pattern is plausibly consistent with prior observations that larger models tend to exhibit improved robustness under adversarial or distribution-shifted prompting conditions~\citep{ganguli2022predictability}.

\para{Regression analysis: Accuracy, Entropy, and Perplexity.}
~\autoref{tab:datasetwise_accuracy_entropy_perplexity_regression} presents mixed-effects regression results examining the relationship between toxicity, prompt features, and model behavior across \arc{}, \gsm{}, and \mml{}.
Across all datasets, toxicity score is negatively associated with accuracy and positively associated with entropy, whereas its association with perplexity is negative (significant on \arc{} and \gsm{}, not significant on \mml{}). 
With toxicity increase, accuracy decreases while entropy increases, indicating such perturbations make model predictions less stable. 

Question length shows a dataset-dependent relationship with degraded performance. Longer questions are associated with lower accuracy and higher entropy and perplexity on \gsm{} and \mml{}, whereas on \arc{} the effects are significant but negligible in magnitude ($|\beta| \le 0.009$), suggesting that increasing input complexity amplifies predictive uncertainty~\cite{dziri2023faith}.
Among the prompt-level features, answer rarity exhibits the strongest negative association with accuracy while its association with entropy and perplexity is mixed (\textit{+} on \arc{}, \textit{-} on \gsm{}), suggesting that rare outputs are not consistently associated with greater uncertainty during generation. 

The regression analysis also reveals differences across model families. Gemma-2-2B and LLaMA-3.2-1B show negative accuracy coefficients alongside strong positive entropy coefficients across datasets. 
These coefficients reflect lower overall accuracy and higher entropy relative to the reference model (Qwen2.5-1.5B-Instruct), consistent with their lower accuracy in ~\autoref{tab:model_accuracy_comparison}.

\section{RQ2: Prompt Perturbation-Induced Mechanistic Changes in LLMs}

\begin{table*}[t]
\centering
\footnotesize
\renewcommand*{\arraystretch}{0.9}
\resizebox{\textwidth}{!}{
\begin{tabular}{lcccccccccccccccc}

& \multicolumn{8}{c}{\textbf{Core Nodes}} 
& \multicolumn{8}{c}{\textbf{Variant Nodes}} \\
\cmidrule(r){2-9} 
\cmidrule(l){10-17}

\textbf{Prompt} 
& \textbf{Act.} & \textbf{$\Delta$\%} & \textbf{$d$} & \textbf{$t$}
& \textbf{Infl.} & \textbf{$\Delta$\%} & \textbf{$d$} & \textbf{$t$}
& \textbf{Act.} & \textbf{$\Delta$\%} & \textbf{$d$} & \textbf{$t$}
& \textbf{Infl.} & \textbf{$\Delta$\%} & \textbf{$d$} & \textbf{$t$} \\

\midrule

Baseline 
& 0.280 & \gradcell{0.00} & --- & ---
& 3.404 & \gradcell{0.00} & --- & ---
& 0.051 & \gradcell{0.00} & --- & ---
& 0.856 & \gradcell{0.00} & --- & --- \\

Polite 
& 0.168 & \gradcell{-40.0} & -0.039 & -1.68
& 2.483 & \gradcell{-27.1} & -0.035 & -1.42
& 0.106 & \gradcell{+107.8} & 0.069 & 2.85**
& 0.999 & \gradcell{+16.7} & 0.026 & 2.08* \\

Random 
& 0.190 & \gradcell{-32.1} & -0.032 & -1.29
& 2.519 & \gradcell{-26.0} & -0.028 & -1.18
& 0.130 & \gradcell{+154.9} & 0.092 & 3.58***
& 1.324 & \gradcell{+54.7} & 0.056 & 3.02** \\

Low Toxic 
& 0.191 & \gradcell{-31.8} & -0.034 & -1.36
& 2.689 & \gradcell{-21.0} & -0.027 & -1.11
& 0.109 & \gradcell{+113.7} & 0.061 & 2.49*
& 1.037 & \gradcell{+21.1} & 0.027 & 2.01* \\

Med. Toxic 
& 0.162 & \gradcell{-42.1} & -0.052 & -2.27*
& 2.062 & \gradcell{-39.4} & -0.051 & -2.11*
& 0.125 & \gradcell{+145.1} & 0.107 & 4.08***
& 1.088 & \gradcell{+27.1} & 0.041 & 2.84** \\

High Toxic 
& 0.172 & \gradcell{-38.6} & -0.047 & -2.01*
& 2.580 & \gradcell{-24.2} & -0.033 & -1.96*
& 0.134 & \gradcell{+162.8} & 0.104 & 4.55***
& 1.458 & \gradcell{+70.3} & 0.071 & 3.31*** \\

\end{tabular}}
\caption{Node-level Activation (Act.) and Influence (Infl.) changes on \arc{}. $\Delta$\% denotes percentage change relative to the baseline condition (\hlpos{teal}: positive; \hlneg{pink}: negative; shading indicates magnitude), along with Cohen's $d$ and $t$-tests (* $p$<0.05, ** $p$<0.01, *** $p$<0.001). Similar analysis for \gsm{} (\autoref{tab:node_activation_influence_gsm8k}) and \mml{} (\autoref{tab:node_activation_influence_mmlu}).}
\label{tab:node_activation_influence_arceasy}
\vspace{-5pt}
\end{table*}

\begin{table}[t]
\centering
\scriptsize
\renewcommand*{\arraystretch}{0.85}
\setlength{\tabcolsep}{2pt}
\resizebox{\columnwidth}{!}{
\begin{tabular}{lllllll}
& \multicolumn{2}{c}{\textbf{\arc{}}} 
& \multicolumn{2}{c}{\textbf{\gsm{}}} 
& \multicolumn{2}{c}{\textbf{\mml{}}} \\
\cmidrule(r){2-3} \cmidrule(r){4-5} \cmidrule(l){6-7}

\textbf{Independent V.} 
& \textbf{Act.} & \textbf{Infl.}
& \textbf{Act.} & \textbf{Infl.}
& \textbf{Act.} & \textbf{Infl.} \\

\midrule

Node Type: Variant
& -0.604*** & -0.714***
& -0.137*** & -0.381***
& -0.112*** & -0.351*** \\

Toxicity Score & 0.200* & 0.158**
& 0.008 & -0.036*
& -0.091* & 0.173** \\

Tox. Score X Variant
& 0.367*** & 0.310***
& 0.011*** & 0.158***
& 0.030*** & 0.042*** \\

Entropy
& 0.141* & 0.029
& -0.061* & -0.110**
& 0.359 & -0.448* \\

Perplexity
& -0.050* & -0.133*
& 0.030** & 0.074**
& -0.156* & 0.544* \\

\hdashline
\rowcollight $R^2$ (marginal)
& 0.436 & 0.580
& 0.028 & 0.169
& 0.061 & 0.173 \\

\rowcollight $R^2$ (conditional)
& 0.592 & 0.734
& 0.143 & 0.312
& 0.194 & 0.341 \\

\end{tabular}}
\caption{Mixed-effects regression coefficients for node activation (Act.) and influence (Infl.) (* $p$<0.05, ** $p$<0.01, *** $p$<0.001).}
\label{tab:datasetwise_activation_influence_regression}
\vspace{-10pt}
\end{table}















\subsection{RQ2: Methodology}

\para{Attribution Graph Generation.}
To gain mechanistic insight into how prompt perturbations influence model outcomes, we construct attribution graphs that trace the internal flow of information from input tokens to output logits. 
Our approach builds on recent advances in circuit-level interpretability, particularly attribution graphs and transcoders that approximate model internals in a more interpretable feature space~\cite{anthropic2024circuits,olah2020circuits}. 
These methods aim to identify structured computational pathways (``circuits'') responsible for specific model behaviors, enabling analysis beyond input--output correlations.

For three open-source models with publicly available transcoders, we generate attribution graphs using a replacement-model framework integrating pretrained language models with learned transcoders~\citep{anthropic2024circuits}. Specifically, we use the \textit{gemma-scope-transcoders} for Gemma-2-2B~\cite{mntss_gemma_scope_transcoders}, \textit{transcoder-Llama-3.2-1B} for LLaMA-3.2-1B~\cite{mntss_llama32_transcoder}, and \textit{qwen3-1.7b-transcoders-lowl0} for Qwen3-1.7B~\cite{whanna_qwen_transcoders}. 
These transcoders project high-dimensional activations into sparse and interpretable feature representations, enabling attribution at the level of latent computational features rather than individual neurons.    
To check faithfulness, we compared each transcoder replacement model with its original model. Transcoder accuracy stays within 4.7 points of the original in every prompt condition, with a mean absolute difference of 1.8 points (\autoref{tab:original_transcoder_feature_comparison}).

Given a prompt \Num{p_{i,j}} constructed using perturbation token \Num{w_j}, we first obtain the model prediction deterministically and then compute an attribution graph representing how internal features contribute to the final output logits. Formally, for a model $f_\theta$ and prompt \Num{p_{i,j}}: $\mathtt{\hat{y}_{i,j} = f_\theta(p_{i,j})}$, where $\hat{y}_{i,j}$ denotes the generated prediction. The resulting attribution graph is represented as: \Num{G_{i,j} = (V_{i,j}, E_{i,j})}, where $V_{i,j}$ denotes internal feature nodes and $E_{i,j}$ represents attributed influence between nodes.

For each feature node $v \in V_{i,j}$, we focus on two complementary quantities: \textbf{activation} and \textbf{influence}. Let $a_v$ denote the activation magnitude of node $v$, and let $I_v$ denote its attributed influence on the final prediction logit. The overall attribution graph can therefore be represented as: \Num{G_{i,j} = \{(v, a_v, I_v) \mid v \in V_{i,j}\}}, where $a_v$ captures how strongly a feature is activated under a given prompt, while $I_v$ captures how much that feature contributes to the final prediction. 
Influence values are computed through attribution over output logits, enabling us to quantify which internal features most strongly affect model decisions~\cite{olah2020circuits,sundararajan2017axiomatic}. 

\para{Identifying Core and Variant Nodes.}
We separate \textit{core} nodes, which support stable reasoning, from \textit{variant} nodes, which are sensitive to prompt perturbations, based on their activation variance across perturbations. This follows prior work linking stable features to invariant reasoning and variable features to context-sensitive processing~\cite{olah2020circuits,anthropic2024circuits}.
For each question $q_i$, we generate attribution graphs for all $K$ perturbations, normalize activations within each graph to remove scale effects, and compute $\mathrm{Var}(v) = \mathrm{Var}(\tilde{a}_v^{(1)}, \dots, \tilde{a}_v^{(K)})$ for each feature node $v$, where $\tilde{a}_v^{(k)}$ is its normalized activation under perturbation $k$.
Nodes below the $25^{\text{th}}$ percentile are \textit{core}, and those above the $75^{\text{th}}$ are \textit{variant}.
Relative to random nodes (\autoref{tab:toxic_node_changes}), mean absolute percentage change in activation is lowest for core nodes (41.20\%), then random (121.85\%), then variant (202.49\%).
Influence follows the same order (40.10\%, 44.63\%, 49.15\%) with smaller gaps, except on \mml{}, where variant nodes change less than core nodes.
Since selection uses activation alone, this ordering suggests the two node types play distinct computational roles.

\para{Regression Modeling of Activation and Influence.}
To further characterize how perturbations affect internal computation, we conduct linear mixed-effects regression.
For each attribution graph, we define two dependent variables: 1) \textit{activation}, representing the average activation magnitude of a node, and 2) \textit{influence}, representing the attributed contribution of that node to the final prediction. 
For independent variables, we use the node type (core or variant) and the toxicity score of the perturbation word. 
We also include an interaction term to capture whether increasing toxicity disproportionately affects variant nodes relative to core reasoning nodes. 
Furthermore, we include the model's entropy and perplexity to capture generation uncertainty and predictive confidence. 




\subsection{RQ2: Results}


\para{Changes in Activation \& Influence.}
\autoref{tab:node_activation_influence_arceasy} summarizes activation and influence changes across core and variant nodes relative to the baseline condition on \arc{}. 
Core nodes show a consistent decrease in activation and influence, which is not significant for polite, random, and low-toxic prompts but becomes significant for medium- and high-toxic prompts.

We observe that polite perturbations increase activation (by 107.84\%) and influence (by 16.71\%) among variant nodes, relative to the baseline condition. 
Also, random perturbations substantially increase variant-node activation (by 154.90\%) and influence (by 54.67\%) relative to the baseline. 
Among toxic perturbations, these shifts grow with toxicity. 
Variant-node activation progressively increases across low-toxic (113.73\%), medium-toxic (145.10\%), and high-toxic (162.75\%) conditions, with influence similarly increasing by 21.14\%, 27.10\%, and 70.33\%, respectively. 
Variant-node activation also increases on \gsm{} and \mml{} (Tables~\ref{tab:node_activation_influence_gsm8k} and~\ref{tab:node_activation_influence_mmlu}), while variant-node influence is more mixed on \mml{} (it decreases under random and low-toxic prompts).
These findings suggest that increasingly toxic prompts progressively recruit perturbation-sensitive computational pathways and make them more influential in the final prediction process.
Finally, we observe that increases in variant-node activation and influence occur while core-node activation and influence decrease relative to the baseline (by 21--42\% on \arc{}, significantly only for medium- and high-toxic prompts).
Perturbations may redirect computation toward perturbation-sensitive features and away from stable reasoning pathways, which may help explain the accuracy degradation in RQ1.


\para{Top Activated Features.}
\autoref{tab:top_activated_nodes_questions} lists the top activated feature per perturbation for one question per dataset.
These features mirror the perturbation: ``please''/``kindly'' for polite prompts, insults and profanity (``idiot,'' ``fuck'') for toxic prompts, and dedicated features (``banana,'' ``quantum'') for random prompts. Even neutral insertions therefore recruit separate pathways, consistent with their effect on variant nodes.
The features are largely question-independent, often topping a perturbation across all three questions (e.g., L18/F10625 for high-toxic prompts in Gemma-2-2B).
High-toxic activations (10.07--12.60) always exceed low-toxic ones (2.93--5.92), yet across perturbation types magnitude does not track toxicity: polite and random features reach up to 16.45 and 13.17, in line with their accuracy drops in RQ1.
Most features lie in middle-to-late layers, and some also respond to unrelated tokens (e.g., ``fra'' in Gemma-2-2B), so perturbation features are not always cleanly disentangled.

\para{Regression analysis of activation and influence.}
~\autoref{tab:datasetwise_activation_influence_regression} shows the mixed effects regression coefficients.
Across datasets, the \textit{variant node type} coefficient is negative for both activation and influence, whereas the interaction term is consistently positive and significant. 
In \arc{}, the interaction coefficient is 0.367 for activation and 0.310 for influence, with smaller but still positive interactions in \gsm{} and \mml{}. 
These interaction effects are more consistent than the direct toxicity score coefficients, whose signs remain mixed across datasets.
Together, these results suggest that increasing toxicity does not uniformly increase activation, but instead selectively amplifies perturbation-sensitive pathways. 
As toxicity increases, variant nodes become more active and influential relative to core nodes, consistent with the node-level trends on \arc{} (\autoref{tab:node_activation_influence_arceasy}). 
Finally, perplexity is significantly associated with activation and influence across all datasets, while entropy shows mostly significant but dataset-dependent effects, indicating that uncertainty-related properties are associated with changes in internal circuits.

\section{Discussion and Conclusion}
\label{section:discussion}

Our results show that even minimal lexical perturbations, whether polite, random, or toxic, alter both the outputs and internal computation of LLMs, reducing accuracy in most models and amplifying perturbation-sensitive circuits over core ones. Increasing toxicity intensifies these effects, further lowering accuracy and raising entropy. We next discuss the implications.

\para{Perturbation Sensitivity and the Role of Toxicity.}
Semantically irrelevant perturbations disrupt both outputs and internal circuits regardless of tone: random words reduce accuracy in all seven models, and even polite words do so in five of seven while amplifying variant nodes. This is consistent with prior evidence that LLMs rely on shallow heuristics and are brittle to minor prompt changes~\cite{niven2019probing,kaushik2020learning,lu2022fantastically,reynolds2021prompt}.
Toxicity intensifies this sensitivity in a graded way: accuracy declines from low- to high-toxic prompts in four of seven models, and higher toxicity is associated with lower accuracy and higher entropy across all datasets.
Toxic phrasing thus poses a growing reliability risk beyond generic lexical noise.

\para{Comparing Core and Perturbation-Sensitive Circuits.}
Our mechanistic analyses reveal a consistent separation between core and perturbation-sensitive circuits. 
Core nodes lose activation and influence under perturbations (significantly for medium- and high-toxic prompts), whereas variant nodes become more active and influential, most clearly and progressively on \arc{}. 
One possible interpretation is that lexical perturbations redirect computation toward perturbation-sensitive pathways, reducing the contribution of stable reasoning circuits responsible for solving the underlying task. 
Similar dynamics have been observed in studies of spurious correlations and shortcut learning, where models over-rely on non-robust features under distribution shifts~\cite{sagawa2020investigation,tu2020empirical}. 

\para{Uncertainty and Internal Representation Shift.}
The entropy increase that accompanies toxicity also links behavior to internal computation: uncertainty is associated with changes in node activation and influence, suggesting that hallucination-like behavior under perturbed, and especially toxic, prompts may stem from unstable intermediate representations and shifting internal circuits, consistent with prior work tying uncertainty and hallucination to representational instability and calibration failures~\cite{desai2021calibration,kadavath2022language}.
Answer rarity also strongly predicts lower accuracy across datasets, suggesting that distributionally sparse answers are especially vulnerable under perturbation. This aligns with prior work showing that long-tail knowledge and infrequent targets remain challenging even for models with strong overall benchmark performance~\cite{mallen2023llm,longpre2024pretraining}.

\para{Prompt Robustness Beyond Adversarial Attacks.}
Our analyses suggest that prompt robustness should not be viewed solely through the lens of jailbreak attacks or explicitly malicious prompts. 
Instead, large language models appear broadly sensitive to auxiliary lexical context that shifts computation away from stable reasoning pathways. 
Prior work on prompt engineering and prompt brittleness similarly demonstrates that semantically equivalent prompts can produce substantially different model behaviors~\citep{shi2023large}. 
Our results suggest that these sensitivities may arise from redistribution of computation toward perturbation-sensitive circuits, a shift that becomes more pronounced as toxicity increases.

\para{Mechanistic Interpretability for Safety Evaluation.}
Our study highlights the value of mechanistic interpretability for studying model robustness and safety. 
While behavioral evaluations show that lexical perturbations reduce accuracy, attribution graphs reveal how internal computation shifts in response. 
In particular, the increasing influence of perturbation-sensitive circuits under toxic prompts provides a mechanistic explanation for degraded model reliability. 
These insights may support future intervention, steering, and alignment methods aimed at suppressing perturbation-sensitive, and particularly toxicity-sensitive, computational pathways~\citep{meng2022locating,chan2022causal}.


\section{Ethical Considerations}

This work studies how lexical perturbations influence the behavior and internal circuits of LLMs using publicly available benchmark datasets (\arc{}, \gsm{}, and \mml{}) and pretrained models. 
The study does not involve human participants or sensitive personal data and therefore did not require institutional ethics approval.
Our experiments include toxic lexical perturbations solely for controlled robustness analysis. 
These perturbations were manually curated and restricted to short lexical modifiers to isolate the effect of toxicity while preserving the semantic meaning of the underlying question. The goal of this work is not to generate harmful content, but to better understand how toxic and non-semantic perturbations affect model reliability and hallucination behavior.

Our findings suggest that toxic prompts can shift model computation toward perturbation-sensitive pathways associated with increased uncertainty and degraded performance. Such behavior may have implications for deploying language models in high-stakes settings where prompt sensitivity can affect reliability.
Finally, our mechanistic analyses rely on attribution graphs and sparse transcoders, which provide only approximate interpretations of internal computation. Therefore, we caution against over-interpreting LLM circuits or treating interpretability results as definitive causal explanations.


\section{Limitations and Future Work}

Our work has limitations which also suggest interesting future directions. 
First, our experiments focus primarily on short-form reasoning tasks with constrained decoding, which may not fully capture open-ended conversational generation (see \autoref{tab:model_mean_comparison_free_form} for a free-form evaluation).
Second, attribution graphs and transcoders provide only an approximate view of internal computation and may not recover all relevant circuits. Third, toxicity itself remains socially contextual and difficult to operationalize using a single scalar score.
Future work can investigate whether similar perturbation-sensitive pathways emerge in multi-turn dialogue systems, chain-of-thought reasoning, and agentic planning settings. 
Extending these analyses to larger frontier models and multilingual contexts may also help determine whether the observed circuit-level dynamics generalize across architectures and deployment settings. Finally, future research can also explore targeted interventions that suppress perturbation-sensitive pathways or reinforce stable reasoning circuits under adversarial lexical conditions. 

\section{AI Involvement Disclosure}
AI-assisted language editing was used exclusively to improve grammar and readability. The study design, analyses, interpretations, and experiments were conducted fully by the authors. 



\bibliography{references}

\appendix
\section{Appendix}
\makeatletter
\setlength{\@dblfptop}{0pt}                
\setlength{\@dblfpsep}{12pt plus 2pt}
\setlength{\@dblfpbot}{0pt plus 1fil}
\setlength{\@fptop}{0pt}                   
\setlength{\@fpsep}{12pt plus 2pt}
\setlength{\@fpbot}{0pt plus 1fil}
\makeatother

\setcounter{table}{0}
\setcounter{figure}{0}
\renewcommand{\thetable}{A\arabic{table}}
\renewcommand{\thefigure}{A\arabic{figure}}


\begin{table}[!b]
\centering
\small
\begin{tabular}{ll}
\toprule
Model & Params \\
\midrule
GPT-5-nano~\cite{singh2025openai} & n/d \\
Gemini-2.5-Flash~\cite{comanici2025gemini} & n/d \\
Gemma-2-2B~\cite{gemmateam2024gemma2} & 2.6B \\
Qwen2.5-1.5B-Instruct~\cite{qwen2024qwen25} & 1.5B \\
Qwen2.5-7B-Instruct~\cite{qwen2024qwen25} & 7.6B \\
LLaMA-3.2-1B~\cite{grattafiori2024llama3} & 1.2B \\
LLaMA-3-8B-Instruct~\cite{grattafiori2024llama3} & 8.0B \\
Qwen3-1.7B~\cite{yang2025qwen3} & 1.7B \\
\bottomrule
\end{tabular}
\caption{Models used in this work (n/d = not disclosed).}
\label{tab:model_details}
\end{table}

\begin{table}[!b]
\centering
\footnotesize
\renewcommand*{\arraystretch}{0.9}
\begin{tabular}{lccc}
& \textbf{Core} & \textbf{Variant} & \textbf{Random} \\
\midrule
\textbf{Activation}
& 41.20
& 202.49
& 121.85 \\

\textbf{Influence}
& 40.10
& 49.15
& 44.63 \\
\bottomrule
\end{tabular}
\caption{Mean absolute percentage changes in node activation and influence across perturbation conditions, averaged across ARC-Easy, GSM8K, and MMLU.}
\label{tab:toxic_node_changes}
\vspace{-5pt}
\end{table} 

\begin{table*}[!tp]
\centering
\footnotesize
\renewcommand*{\arraystretch}{0.9}
\setlength{\tabcolsep}{2pt}
\resizebox{\textwidth}{!}{%
\begin{tabular}{lcccccccccccccccccccccccccccc}
& \multicolumn{4}{c}{\textbf{GPT-5-nano}}
& \multicolumn{4}{c}{\textbf{Gemini-2.5-Flash}}
& \multicolumn{4}{c}{\textbf{Gemma-2-2B}}
& \multicolumn{4}{c}{\textbf{Qwen2.5-1.5B}}
& \multicolumn{4}{c}{\textbf{LLaMA-3.2-1B}}
& \multicolumn{4}{c}{\textbf{LLaMA-3-8B-Instruct}}
& \multicolumn{4}{c}{\textbf{Qwen2.5-7B-Instruct}} \\

\cmidrule(r){2-5}
\cmidrule(lr){6-9}
\cmidrule(lr){10-13}
\cmidrule(lr){14-17}
\cmidrule(lr){18-21}
\cmidrule(lr){22-25}
\cmidrule(l){26-29}

\textbf{Bin}
& \textbf{Acc.} & \textbf{$\Delta$\%} & \textbf{$d$} & \textbf{$t$}
& \textbf{Acc.} & \textbf{$\Delta$\%} & \textbf{$d$} & \textbf{$t$}
& \textbf{Acc.} & \textbf{$\Delta$\%} & \textbf{$d$} & \textbf{$t$}
& \textbf{Acc.} & \textbf{$\Delta$\%} & \textbf{$d$} & \textbf{$t$}
& \textbf{Acc.} & \textbf{$\Delta$\%} & \textbf{$d$} & \textbf{$t$}
& \textbf{Acc.} & \textbf{$\Delta$\%} & \textbf{$d$} & \textbf{$t$}
& \textbf{Acc.} & \textbf{$\Delta$\%} & \textbf{$d$} & \textbf{$t$} \\

\midrule

Baseline
& 0.3667 & \gradcell{0.00} & --- & ---
& 0.3617 & \gradcell{0.00} & --- & ---
& 0.3967 & \gradcell{0.00} & --- & ---
& 0.3900 & \gradcell{0.00} & --- & ---
& 0.1967 & \gradcell{0.00} & --- & ---
& 0.3900 & \gradcell{0.00} & --- & ---
& 0.4000 & \gradcell{0.00} & --- & --- \\

Polite
& 0.3583 & \gradcell{-2.27} & -0.0451 & -0.78
& 0.3583 & \gradcell{-0.92} & -0.0204 & -0.35
& 0.3267 & \gradcell{-17.65} & -0.1845 & -3.20**
& 0.3717 & \gradcell{-4.70} & -0.0698 & -1.21
& 0.2267 & \gradcell{+15.25} & 0.0921 & 1.60
& 0.3367 & \gradcell{-13.68} & -0.1938 & -3.36***
& 0.3933 & \gradcell{-1.67} & -0.0333 & -0.58 \\

Random
& 0.3350 & \gradcell{-8.64} & -0.1441 & -2.50*
& 0.3400 & \gradcell{-5.99} & -0.1124 & -1.95
& 0.2583 & \gradcell{-34.87} & -0.3240 & -5.61***
& 0.2517 & \gradcell{-35.47} & -0.3906 & -6.76***
& 0.1917 & \gradcell{-2.54} & -0.0169 & -0.29
& 0.3650 & \gradcell{-6.41} & -0.0823 & -1.43
& 0.3617 & \gradcell{-9.58} & -0.1246 & -2.16* \\

Low Toxic
& 0.3583 & \gradcell{-2.27} & -0.0440 & -0.76
& 0.3683 & \gradcell{+1.84} & 0.0374 & 0.65
& 0.2983 & \gradcell{-24.79} & -0.2458 & -4.26***
& 0.3700 & \gradcell{-5.13} & -0.0627 & -1.09
& 0.2117 & \gradcell{+7.63} & 0.0464 & 0.80
& 0.4033 & \gradcell{+3.42} & 0.0504 & 0.87
& 0.3683 & \gradcell{-7.92} & -0.1441 & -2.50* \\

Medium Toxic
& 0.3517 & \gradcell{-4.09} & -0.1003 & -1.74
& 0.3683 & \gradcell{+1.84} & 0.0408 & 0.71
& 0.3050 & \gradcell{-23.11} & -0.2411 & -4.18***
& 0.3267 & \gradcell{-16.24} & -0.1973 & -3.42***
& 0.2183 & \gradcell{+11.02} & 0.0726 & 1.26
& 0.4000 & \gradcell{+2.56} & 0.0357 & 0.62
& 0.3617 & \gradcell{-9.58} & -0.1617 & -2.80** \\

High Toxic
& 0.3633 & \gradcell{-0.91} & -0.0182 & -0.32
& 0.3483 & \gradcell{-3.69} & -0.0731 & -1.27
& 0.2933 & \gradcell{-26.05} & -0.2612 & -4.52***
& 0.3350 & \gradcell{-14.10} & -0.1712 & -2.97**
& 0.2233 & \gradcell{+13.56} & 0.0908 & 1.57
& 0.3933 & \gradcell{+0.85} & 0.0120 & 0.21
& 0.3567 & \gradcell{-10.83} & -0.1565 & -2.71** \\

\end{tabular}%
}
\caption{Model-wise accuracy (Acc.) on \arc{}, under lexical perturbation conditions relative to the baseline, with $\Delta$\% denoting percentage change in accuracy relative to baseline (\hlpos{teal}: positive; \hlneg{pink}: negative; shading indicates magnitude), Cohen's $d$, and $t$-tests ($*$ $p<0.05$, $**$ $p<0.01$, $***$ $p<0.001$).
}
\label{tab:model_accuracy_comparison_arc_easy}
\vspace{-5pt}
\end{table*}

\begin{table*}[!tp]
\centering
\footnotesize
\renewcommand*{\arraystretch}{0.9}
\setlength{\tabcolsep}{2pt}
\resizebox{\textwidth}{!}{%
\begin{tabular}{lcccccccccccccccccccccccccccc}
& \multicolumn{4}{c}{\textbf{GPT-5-nano}}
& \multicolumn{4}{c}{\textbf{Gemini-2.5-Flash}}
& \multicolumn{4}{c}{\textbf{Gemma-2-2B}}
& \multicolumn{4}{c}{\textbf{Qwen2.5-1.5B}}
& \multicolumn{4}{c}{\textbf{LLaMA-3.2-1B}}
& \multicolumn{4}{c}{\textbf{LLaMA-3-8B-Instruct}}
& \multicolumn{4}{c}{\textbf{Qwen2.5-7B-Instruct}} \\

\cmidrule(r){2-5}
\cmidrule(lr){6-9}
\cmidrule(lr){10-13}
\cmidrule(lr){14-17}
\cmidrule(lr){18-21}
\cmidrule(lr){22-25}
\cmidrule(l){26-29}

\textbf{Bin}
& \textbf{Acc.} & \textbf{$\Delta$\%} & \textbf{$d$} & \textbf{$t$}
& \textbf{Acc.} & \textbf{$\Delta$\%} & \textbf{$d$} & \textbf{$t$}
& \textbf{Acc.} & \textbf{$\Delta$\%} & \textbf{$d$} & \textbf{$t$}
& \textbf{Acc.} & \textbf{$\Delta$\%} & \textbf{$d$} & \textbf{$t$}
& \textbf{Acc.} & \textbf{$\Delta$\%} & \textbf{$d$} & \textbf{$t$}
& \textbf{Acc.} & \textbf{$\Delta$\%} & \textbf{$d$} & \textbf{$t$}
& \textbf{Acc.} & \textbf{$\Delta$\%} & \textbf{$d$} & \textbf{$t$} \\

\midrule

Baseline
& 0.9483 & \gradcell{0.00} & --- & ---
& 0.8200 & \gradcell{0.00} & --- & ---
& 0.0133 & \gradcell{0.00} & --- & ---
& 0.0567 & \gradcell{0.00} & --- & ---
& 0.0267 & \gradcell{0.00} & --- & ---
& 0.1067 & \gradcell{0.00} & --- & ---
& 0.1933 & \gradcell{0.00} & --- & --- \\

Polite
& 0.9450 & \gradcell{-0.35} & -0.0235 & -0.41
& 0.8517 & \gradcell{+3.86} & 0.0831 & 1.44
& 0.0050 & \gradcell{-62.50} & -0.0965 & -1.67
& 0.0633 & \gradcell{+11.76} & 0.0421 & 0.73
& 0.0483 & \gradcell{+81.25} & 0.1150 & 1.99*
& 0.0983 & \gradcell{-7.81} & -0.0430 & -0.74
& 0.1767 & \gradcell{-8.62} & -0.0609 & -1.05 \\

Random
& 0.9550 & \gradcell{+0.70} & 0.0544 & 0.94
& 0.5300 & \gradcell{-35.37} & -0.6541 & -11.33***
& 0.0150 & \gradcell{+12.50} & 0.0192 & 0.33
& 0.0817 & \gradcell{+44.12} & 0.1133 & 1.96
& 0.0383 & \gradcell{+43.75} & 0.0884 & 1.53
& 0.1183 & \gradcell{+10.94} & 0.0494 & 0.85
& 0.2000 & \gradcell{+3.45} & 0.0255 & 0.44 \\

Low Toxic
& 0.9517 & \gradcell{+0.35} & 0.0258 & 0.45
& 0.8217 & \gradcell{+0.20} & 0.0047 & 0.08
& 0.0133 & \gradcell{0.00} & 0.0000 & 0.00
& 0.0650 & \gradcell{+14.71} & 0.0421 & 0.73
& 0.0467 & \gradcell{+75.00} & 0.1129 & 1.96
& 0.1083 & \gradcell{+1.56} & 0.0082 & 0.14
& 0.1283 & \gradcell{-33.62} & -0.1848 & -3.20** \\

Medium Toxic
& 0.9500 & \gradcell{+0.18} & 0.0140 & 0.24
& 0.7533 & \gradcell{-8.13} & -0.1765 & -3.06**
& 0.0117 & \gradcell{-12.50} & -0.0192 & -0.33
& 0.0650 & \gradcell{+14.71} & 0.0502 & 0.87
& 0.0417 & \gradcell{+56.25} & 0.0880 & 1.52
& 0.0983 & \gradcell{-7.81} & -0.0430 & -0.74
& 0.2250 & \gradcell{+16.38} & 0.1131 & 1.96 \\

High Toxic
& 0.9500 & \gradcell{+0.18} & 0.0149 & 0.26
& 0.5017 & \gradcell{-38.82} & -0.7644 & -13.24***
& 0.0133 & \gradcell{0.00} & 0.0000 & 0.00
& 0.0667 & \gradcell{+17.65} & 0.0548 & 0.95
& 0.0417 & \gradcell{+56.25} & 0.0968 & 1.68
& 0.1083 & \gradcell{+1.56} & 0.0081 & 0.14
& 0.1950 & \gradcell{+0.86} & 0.0066 & 0.11 \\

\end{tabular}%
}
\caption{Model-wise accuracy (Acc.) on \gsm{}, under lexical perturbation conditions relative to the baseline, with $\Delta$\% denoting percentage change in accuracy relative to baseline (\hlpos{teal}: positive; \hlneg{pink}: negative; shading indicates magnitude), Cohen's $d$, and $t$-tests ($*$ $p<0.05$, $**$ $p<0.01$, $***$ $p<0.001$).
}
\label{tab:model_accuracy_comparison_gsm8k}
\vspace{-5pt}
\end{table*}

\begin{table*}[!tp]
\centering
\footnotesize
\renewcommand*{\arraystretch}{0.9}
\setlength{\tabcolsep}{2pt}
\resizebox{\textwidth}{!}{%
\begin{tabular}{lcccccccccccccccccccccccccccc}
& \multicolumn{4}{c}{\textbf{GPT-5-nano}}
& \multicolumn{4}{c}{\textbf{Gemini-2.5-Flash}}
& \multicolumn{4}{c}{\textbf{Gemma-2-2B}}
& \multicolumn{4}{c}{\textbf{Qwen2.5-1.5B}}
& \multicolumn{4}{c}{\textbf{LLaMA-3.2-1B}}
& \multicolumn{4}{c}{\textbf{LLaMA-3-8B-Instruct}}
& \multicolumn{4}{c}{\textbf{Qwen2.5-7B-Instruct}} \\

\cmidrule(r){2-5}
\cmidrule(lr){6-9}
\cmidrule(lr){10-13}
\cmidrule(lr){14-17}
\cmidrule(lr){18-21}
\cmidrule(lr){22-25}
\cmidrule(l){26-29}

\textbf{Bin}
& \textbf{Acc.} & \textbf{$\Delta$\%} & \textbf{$d$} & \textbf{$t$}
& \textbf{Acc.} & \textbf{$\Delta$\%} & \textbf{$d$} & \textbf{$t$}
& \textbf{Acc.} & \textbf{$\Delta$\%} & \textbf{$d$} & \textbf{$t$}
& \textbf{Acc.} & \textbf{$\Delta$\%} & \textbf{$d$} & \textbf{$t$}
& \textbf{Acc.} & \textbf{$\Delta$\%} & \textbf{$d$} & \textbf{$t$}
& \textbf{Acc.} & \textbf{$\Delta$\%} & \textbf{$d$} & \textbf{$t$}
& \textbf{Acc.} & \textbf{$\Delta$\%} & \textbf{$d$} & \textbf{$t$} \\

\midrule

Baseline
& 0.6267 & \gradcell{0.00} & --- & ---
& 0.5900 & \gradcell{0.00} & --- & ---
& 0.3233 & \gradcell{0.00} & --- & ---
& 0.3333 & \gradcell{0.00} & --- & ---
& 0.1100 & \gradcell{0.00} & --- & ---
& 0.3633 & \gradcell{0.00} & --- & ---
& 0.4233 & \gradcell{0.00} & --- & --- \\

Polite
& 0.5950 & \gradcell{-5.05} & -0.1242 & -2.15*
& 0.5883 & \gradcell{-0.28} & -0.0062 & -0.11
& 0.2650 & \gradcell{-18.04} & -0.1989 & -3.44***
& 0.3100 & \gradcell{-7.00} & -0.0863 & -1.50
& 0.0783 & \gradcell{-28.79} & -0.1618 & -2.80**
& 0.3033 & \gradcell{-16.51} & -0.1881 & -3.26**
& 0.4000 & \gradcell{-5.51} & -0.1226 & -2.12* \\

Random
& 0.5183 & \gradcell{-17.29} & -0.3225 & -5.59***
& 0.5067 & \gradcell{-14.12} & -0.2239 & -3.88***
& 0.1983 & \gradcell{-38.66} & -0.3503 & -6.07***
& 0.2517 & \gradcell{-24.50} & -0.2240 & -3.88***
& 0.0700 & \gradcell{-36.36} & -0.1785 & -3.09**
& 0.2683 & \gradcell{-26.15} & -0.2564 & -4.44***
& 0.3517 & \gradcell{-16.93} & -0.2059 & -3.57*** \\

Low Toxic
& 0.5783 & \gradcell{-7.71} & -0.1780 & -3.08**
& 0.5967 & \gradcell{+1.13} & 0.0212 & 0.37
& 0.1667 & \gradcell{-48.45} & -0.3884 & -6.73***
& 0.2983 & \gradcell{-10.50} & -0.1010 & -1.75
& 0.0867 & \gradcell{-21.21} & -0.1105 & -1.91
& 0.3217 & \gradcell{-11.47} & -0.1434 & -2.48*
& 0.3967 & \gradcell{-6.30} & -0.0945 & -1.64 \\

Medium Toxic
& 0.5717 & \gradcell{-8.78} & -0.2012 & -3.48***
& 0.5667 & \gradcell{-3.95} & -0.0727 & -1.26
& 0.1917 & \gradcell{-40.72} & -0.3532 & -6.12***
& 0.2800 & \gradcell{-16.00} & -0.1534 & -2.66**
& 0.0883 & \gradcell{-19.70} & -0.0933 & -1.62
& 0.2900 & \gradcell{-20.18} & -0.2117 & -3.67***
& 0.3850 & \gradcell{-9.06} & -0.1292 & -2.24* \\

High Toxic
& 0.5467 & \gradcell{-12.77} & -0.2709 & -4.69***
& 0.5583 & \gradcell{-5.37} & -0.1001 & -1.73
& 0.1617 & \gradcell{-50.00} & -0.4398 & -7.62***
& 0.2600 & \gradcell{-22.00} & -0.2074 & -3.59***
& 0.0900 & \gradcell{-18.18} & -0.0964 & -1.67
& 0.2817 & \gradcell{-22.48} & -0.2449 & -4.24***
& 0.3733 & \gradcell{-11.81} & -0.1499 & -2.60** \\

\end{tabular}%
}
\caption{Model-wise accuracy (Acc.) on \mml{}, under lexical perturbation conditions relative to the baseline, with $\Delta$\% denoting percentage change in accuracy relative to baseline (\hlpos{teal}: positive; \hlneg{pink}: negative; shading indicates magnitude), Cohen's $d$, and $t$-tests ($*$ $p<0.05$, $**$ $p<0.01$, $***$ $p<0.001$).
}
\label{tab:model_accuracy_comparison_mmlu}
\vspace{-5pt}
\end{table*}

\begin{table*}[!tp]
\centering
\footnotesize
\renewcommand*{\arraystretch}{0.9}
\setlength{\tabcolsep}{2pt}
\resizebox{\textwidth}{!}{%
\begin{tabular}{lcccccccccccccccccccccccccccc}
& \multicolumn{4}{c}{\textbf{GPT-5-nano}}
& \multicolumn{4}{c}{\textbf{Gemini-2.5-Flash}}
& \multicolumn{4}{c}{\textbf{Gemma-2-2B}}
& \multicolumn{4}{c}{\textbf{Qwen2.5-1.5B}}
& \multicolumn{4}{c}{\textbf{LLaMA-3.2-1B}}
& \multicolumn{4}{c}{\textbf{LLaMA-3-8B-Instruct}}
& \multicolumn{4}{c}{\textbf{Qwen2.5-7B-Instruct}} \\

\cmidrule(r){2-5}
\cmidrule(lr){6-9}
\cmidrule(lr){10-13}
\cmidrule(lr){14-17}
\cmidrule(lr){18-21}
\cmidrule(lr){22-25}
\cmidrule(l){26-29}

\textbf{Bin}
& \textbf{Acc.} & \textbf{$\Delta$\%} & \textbf{$d$} & \textbf{$t$}
& \textbf{Acc.} & \textbf{$\Delta$\%} & \textbf{$d$} & \textbf{$t$}
& \textbf{Acc.} & \textbf{$\Delta$\%} & \textbf{$d$} & \textbf{$t$}
& \textbf{Acc.} & \textbf{$\Delta$\%} & \textbf{$d$} & \textbf{$t$}
& \textbf{Acc.} & \textbf{$\Delta$\%} & \textbf{$d$} & \textbf{$t$}
& \textbf{Acc.} & \textbf{$\Delta$\%} & \textbf{$d$} & \textbf{$t$}
& \textbf{Acc.} & \textbf{$\Delta$\%} & \textbf{$d$} & \textbf{$t$} \\

\midrule

Baseline
& 0.7622 & \gradcell{0.00} & --- & ---
& 0.7944 & \gradcell{0.00} & --- & ---
& 0.1333 & \gradcell{0.00} & --- & ---
& 0.3078 & \gradcell{0.00} & --- & ---
& 0.1800 & \gradcell{0.00} & --- & ---
& 0.3911 & \gradcell{0.00} & --- & ---
& 0.3822 & \gradcell{0.00} & --- & --- \\

Polite
& 0.7578 & \gradcell{-0.58} & -0.019 & -0.56
& 0.8044 & \gradcell{+1.26} & 0.048 & 1.43
& 0.1550 & \gradcell{+16.25} & 0.057 & 1.72
& 0.3006 & \gradcell{-2.35} & -0.019 & -0.57
& 0.1794 & \gradcell{-0.31} & -0.001 & -0.04
& 0.3800 & \gradcell{-2.84} & -0.028 & -0.84
& 0.3861 & \gradcell{+1.02} & 0.013 & 0.40 \\

Random
& 0.7461 & \gradcell{-2.11} & -0.063 & -1.89
& 0.7772 & \gradcell{-2.17} & -0.071 & -2.13*
& 0.1106 & \gradcell{-17.08} & -0.061 & -1.82
& 0.3044 & \gradcell{-1.08} & -0.009 & -0.26
& 0.1783 & \gradcell{-0.93} & -0.004 & -0.13
& 0.3833 & \gradcell{-1.99} & -0.020 & -0.59
& 0.3433 & \gradcell{-10.17} & -0.120 & -3.61*** \\

Low Toxic
& 0.7533 & \gradcell{-1.17} & -0.035 & -1.05
& 0.7933 & \gradcell{-0.14} & -0.005 & -0.15
& 0.1333 & \gradcell{0.00} & 0.000 & 0.00
& 0.3100 & \gradcell{+0.72} & 0.006 & 0.18
& 0.1917 & \gradcell{+6.48} & 0.029 & 0.87
& 0.3900 & \gradcell{-0.28} & -0.003 & -0.08
& 0.3633 & \gradcell{-4.94} & -0.061 & -1.83 \\

Medium Toxic
& 0.7650 & \gradcell{+0.36} & 0.011 & 0.32
& 0.7883 & \gradcell{-0.77} & -0.028 & -0.84
& 0.1339 & \gradcell{+0.42} & 0.001 & 0.04
& 0.2894 & \gradcell{-5.96} & -0.049 & -1.46
& 0.1861 & \gradcell{+3.40} & 0.016 & 0.47
& 0.4017 & \gradcell{+2.70} & 0.025 & 0.76
& 0.3572 & \gradcell{-6.54} & -0.080 & -2.41* \\

High Toxic
& 0.7528 & \gradcell{-1.24} & -0.036 & -1.08
& 0.7750 & \gradcell{-2.45} & -0.087 & -2.61**
& 0.1206 & \gradcell{-9.58} & -0.034 & -1.01
& 0.2911 & \gradcell{-5.42} & -0.046 & -1.37
& 0.1639 & \gradcell{-8.95} & -0.041 & -1.24
& 0.3894 & \gradcell{-0.43} & -0.004 & -0.13
& 0.3789 & \gradcell{-0.87} & -0.011 & -0.33 \\

\end{tabular}%
}
\caption{Model-wise mean accuracy values under the evaluated conditions relative to the baseline under \textbf{free form generation}, with $\Delta$\% denoting percentage change relative to baseline (\hlpos{teal}: positive; \hlneg{pink}: negative; shading indicates magnitude), Cohen's $d$, and $t$-tests (* $p<0.05$, ** $p<0.01$, *** $p<0.001$).}
\label{tab:model_mean_comparison_free_form}
\vspace{-5pt}
\end{table*}

\begin{table*}[!tp]
\centering
\footnotesize
\setlength{\tabcolsep}{2.5pt}
\resizebox{\textwidth}{!}{
\begin{tabular}{lccccccccccccccccccccc}
& \multicolumn{3}{c}{\textbf{GPT-5-nano}}
& \multicolumn{3}{c}{\textbf{Gemini-2.5-Flash}}
& \multicolumn{3}{c}{\textbf{Gemma-2-2B}}
& \multicolumn{3}{c}{\textbf{Qwen2.5-1.5B-Instruct}}
& \multicolumn{3}{c}{\textbf{LLaMA-3.2-1B}}
& \multicolumn{3}{c}{\textbf{LLaMA-3-8B-Instruct}}
& \multicolumn{3}{c}{\textbf{Qwen2.5-7B-Instruct}} \\

\cmidrule(r){2-4}
\cmidrule(lr){5-7}
\cmidrule(lr){8-10}
\cmidrule(lr){11-13}
\cmidrule(lr){14-16}
\cmidrule(lr){17-19}
\cmidrule(l){20-22}

\textbf{Prtbn. Position}
& \textbf{Low} & \textbf{Med.} & \textbf{High}
& \textbf{Low} & \textbf{Med.} & \textbf{High}
& \textbf{Low} & \textbf{Med.} & \textbf{High}
& \textbf{Low} & \textbf{Med.} & \textbf{High}
& \textbf{Low} & \textbf{Med.} & \textbf{High}
& \textbf{Low} & \textbf{Med.} & \textbf{High}
& \textbf{Low} & \textbf{Med.} & \textbf{High} \\

\midrule

\textbf{Append}
& \gradcell{-2.75} & \gradcell{-3.52} & \gradcell{-4.21}
& \gradcell{+0.85} & \gradcell{-4.70} & \gradcell{-20.51}
& \gradcell{-34.77} & \gradcell{-30.68} & \gradcell{-36.14}
& \gradcell{-5.98} & \gradcell{-13.89} & \gradcell{-15.17}
& \gradcell{+3.50} & \gradcell{+4.50} & \gradcell{+6.50}
& \gradcell{-3.10} & \gradcell{-8.33} & \gradcell{-8.91}
& \gradcell{-12.13} & \gradcell{-4.43} & \gradcell{-9.02} \\

\textbf{Insert Middle}
& \gradcell{+1.19} & \gradcell{-0.30} & \gradcell{-0.74}
& \gradcell{+1.84} & \gradcell{+0.99} & \gradcell{-1.84}
& \gradcell{-11.68} & \gradcell{-14.02} & \gradcell{-16.36}
& \gradcell{-10.34} & \gradcell{-11.21} & \gradcell{-14.66}
& \gradcell{-16.83} & \gradcell{-10.89} & \gradcell{-19.80}
& \gradcell{-9.06} & \gradcell{-9.45} & \gradcell{-11.02}
& \gradcell{-8.67} & \gradcell{-6.33} & \gradcell{-11.00} \\

\textbf{Prepend}
& \gradcell{+0.30} & \gradcell{-0.15} & \gradcell{-0.15}
& \gradcell{+1.66} & \gradcell{-7.47} & \gradcell{-7.19}
& \gradcell{+2.29} & \gradcell{-0.92} & \gradcell{-9.63}
& \gradcell{-1.71} & \gradcell{-0.85} & \gradcell{-2.14}
& \gradcell{+31.63} & \gradcell{+33.67} & \gradcell{+18.37}
& \gradcell{+8.73} & \gradcell{+1.59} & \gradcell{-8.73}
& \gradcell{-1.64} & \gradcell{-14.14} & \gradcell{-20.39} \\

\bottomrule
\end{tabular}}
\caption{Change in accuracy (\%) relative to the baseline across perturbation positions and toxicity levels.}
\label{tab:perturbation_position_toxicity}
\end{table*}

\begin{table*}[!tp]
\centering
\footnotesize
\setlength{\tabcolsep}{2pt}
\resizebox{\textwidth}{!}{%
\begin{tabular}{lccccccc}
\textbf{Dataset}
& \textbf{GPT-5-nano}
& \textbf{Gemini-2.5-Flash}
& \textbf{Gemma-2-2B}
& \textbf{LLaMA-3-8B-Instruct}
& \textbf{LLaMA-3.2-1B}
& \textbf{Qwen2.5-1.5B}
& \textbf{Qwen2.5-7B-Instruct} \\
\midrule

ARC-Easy
& 0.742\%
& 1.424\%
& 0.000\%
& 0.000\%
& 0.000\%
& 0.076\%
& 0.288\% \\

GSM8K
& 0.000\%
& 0.136\%
& 0.030\%
& 0.000\%
& 0.000\%
& 0.023\%
& 0.000\% \\

MMLU
& 1.220\%
& 2.129\%
& 0.008\%
& 0.030\%
& 0.000\%
& 0.038\%
& 0.167\% \\

\bottomrule
\end{tabular}%
}
\caption{Refusal rates across datasets and models. Values are reported as percentages of responses classified as refusals.}
\label{tab:refusal_rates}
\end{table*}

\begin{table*}[!tp]
\centering
\scriptsize
\renewcommand*{\arraystretch}{1.2}
\setlength{\tabcolsep}{3pt}
\newcolumntype{Q}{>{\raggedright\arraybackslash}p{3.7cm}}
\newcolumntype{F}{>{\centering\arraybackslash}p{2.2cm}}
\newcolumntype{H}{>{\centering\arraybackslash}p{2.2cm}}
\newcommand{\fe}[3]{#1\newline{\tiny\color{gray!70!black}#2 / #3}}

\begin{tabular}{QFFFFH}
\textbf{Question}
& \textbf{Polite}
& \textbf{Random}
& \textbf{Low Toxic}
& \textbf{Medium Toxic}
& \textbf{High Toxic} \\
\midrule

\multicolumn{6}{l}{\textbf{Gemma-2-2B}} \\
\cmidrule(r){1-1}

\textit{\arc{}:} A research scientist writes a paper on the initial regrowth of a forest after a fire has damaged the entire ecosystem. Which title would be best for the paper?
& \fe{Please, Pleas, Pease}{L17/F3097}{8.74}
& \fe{Banana, Monkey, banan}{L19/F13694}{7.12}
& \fe{Sad, Bad, Ugly}{L18/F10232}{3.52}
& \fe{Dumb, stup, stupe}{L20/F6570}{4.76}
& \fe{fuck, fucked, fra}{L18/F10625}{10.07} \\

\textit{\gsm{}:} John runs 60 miles a week. He runs 3 days a week. He runs 3 hours the first day and half as much the other two days he runs. How fast does he run?
& \fe{Please, Pleas, Pease}{L17/F3097}{8.60}
& \fe{Banana, Monkey, banan}{L19/F13694}{7.17}
& \fe{tricky, wierd, weird}{L25/F3799}{5.82}
& \fe{dumbass, idiots, Idiot}{L16/F12}{4.49}
& \fe{fuck, fucked, fra}{L18/F10625}{12.60} \\

\textit{\mml{}:} Suppose today is Wednesday. What day of the week will it be $10^{(10^{10})}$ days from now?
& \fe{Please, Pleas, Pease}{L17/F3097}{9.54}
& \fe{Banana, Monkey, banan}{L19/F13694}{8.07}
& \fe{tricky, wierd, weird}{L25/F3799}{5.92}
& \fe{dumbass, idiots, Idiot}{L16/F12}{5.36}
& \fe{fuck, fucked, fra}{L18/F10625}{11.55} \\

\midrule

\multicolumn{6}{l}{\textbf{Llama-3.2-1B}} \\
\cmidrule(r){1-1}

\textit{\arc{}:} A research scientist writes a paper on the initial regrowth of a forest after a fire has damaged the entire ecosystem. Which title would be best for the paper?
& \fe{please, pls, kindly}{L14/F5703}{14.05}
& \fe{Quantum, Quant, quartz}{L14/F26999}{8.38}
& \fe{Angry, Lonely, Naughty}{L15/F56393}{4.33}
& \fe{idiot, stupidity, idiots}{L2/F13069}{4.81}
& \fe{fuck, fucks, Fucked}{L11/F116103}{10.59} \\

\textit{\gsm{}:} John runs 60 miles a week. He runs 3 days a week. He runs 3 hours the first day and half as much the other two days he runs. How fast does he run?
& \fe{please, pls, kindly}{L14/F5703}{14.44}
& \fe{Quantum, Quant, quartz}{L14/F26999}{5.68}
& \fe{Crazy, creepy, Funny}{L12/F49264}{2.93}
& \fe{idiot, stupidity, idiots}{L2/F13069}{4.87}
& \fe{fuck, fucks, Fucked}{L11/F116103}{10.61} \\

\textit{\mml{}:} Suppose today is Wednesday. What day of the week will it be $10^{(10^{10})}$ days from now?
& \fe{please, pls, kindly}{L15/F85491}{16.45}
& \fe{Quantum, Quant, quartz}{L14/F26999}{7.61}
& \fe{Crazy, creepy, Funny}{L12/F49264}{3.51}
& \fe{idiot, stupidity, idiots}{L2/F13069}{4.63}
& \fe{fuck, fucks, Fucked}{L11/F116103}{10.84} \\

\midrule

\multicolumn{6}{l}{\textbf{Qwen3-1.7B}} \\
\cmidrule(r){1-1}

\textit{\arc{}:} A research scientist writes a paper on the initial regrowth of a forest after a fire has damaged the entire ecosystem. Which title would be best for the paper?
& \fe{PLEASE, kindly, MUST}{L25/F88138}{13.17}
& \fe{Banana, bananas, Ban}{L26/F153239}{12.87}
& \fe{silly, funny, stupid}{L19/F51922}{3.48}
& \fe{Husband, idiot, devil}{L23/F36620}{9.97}
& \fe{FUCK, fucks, Faul}{L24/F86509}{10.48} \\

\textit{\gsm{}:} John runs 60 miles a week. He runs 3 days a week. He runs 3 hours the first day and half as much the other two days he runs. How fast does he run?
& \fe{pls, please, plea}{L24/F50485}{10.33}
& \fe{quantum, Fro, Cler}{L22/F44380}{12.14}
& \fe{silly, funny, stupid}{L19/F51922}{3.25}
& \fe{Husband, idiot, devil}{L23/F36620}{8.21}
& \fe{FUCK, fucks, Faul}{L24/F86509}{10.15} \\

\textit{\mml{}:} Suppose today is Wednesday. What day of the week will it be $10^{(10^{10})}$ days from now?
& \fe{pls, please, plea}{L24/F50485}{11.77}
& \fe{quantum, Fro, Cler}{L22/F44380}{13.17}
& \fe{silly, funny, stupid}{L19/F51922}{3.09}
& \fe{Husband, idiot, devil}{L23/F36620}{6.26}
& \fe{FUCK, fucks, Faul}{L24/F86509}{10.92} \\

\bottomrule
\end{tabular}

\caption{Top activated features for each perturbation on one example question per dataset. Each cell lists the feature's top three trigger words, followed by its layer/feature identifier and activation score.}
\label{tab:top_activated_nodes_questions}
\vspace{-5pt}
\end{table*}

\begin{table*}[!tp]
\centering
\footnotesize
\renewcommand*{\arraystretch}{0.9}
\setlength{\tabcolsep}{4pt}
\resizebox{\textwidth}{!}{%
\begin{tabular}{lcccccc}
\textbf{Prompt}
& \textbf{Gemma-2-2B} & \textbf{Gemma-2-2B}
& \textbf{LLaMA-3.2-1B} & \textbf{LLaMA-3.2-1B}
& \textbf{Qwen3-1.7B} & \textbf{Qwen3-1.7B} \\
& \textbf{(Original)} & \textbf{(Transcoder)}
& \textbf{(Original)} & \textbf{(Transcoder)}
& \textbf{(Original)} & \textbf{(Transcoder)} \\
\midrule

Baseline
& 25.67 & 27.33
& 14.67 & 16.00
& 28.00 & 29.33 \\

Polite
& 25.33 & 23.00
& 12.00 & 13.67
& 23.00 & 21.33 \\

Random
& 15.33 & 16.67
& 10.33 & 12.33
& 28.33 & 23.67 \\

Low Toxic
& 22.67 & 24.33
& 15.33 & 15.00
& 27.33 & 26.00 \\

Medium Toxic
& 20.33 & 22.33
& 12.00 & 13.33
& 21.33 & 24.67 \\

High Toxic
& 21.33 & 21.67
& 12.00 & 15.33
& 25.67 & 26.33 \\

\bottomrule
\end{tabular}%
}
\caption{Comparison of the Accuracy\% across prompts for the original and transcoder representations of Gemma-2-2B, LLaMA-3.2-1B, and Qwen3-1.7B.}
\label{tab:original_transcoder_feature_comparison}
\end{table*}

\begin{table*}[!tp]
\centering
\footnotesize
\renewcommand*{\arraystretch}{0.9}
\resizebox{\textwidth}{!}{
\begin{tabular}{lcccccccccccccccc}

& \multicolumn{8}{c}{\textbf{Core Nodes}}
& \multicolumn{8}{c}{\textbf{Variant Nodes}} \\
\cmidrule(r){2-9}
\cmidrule(l){10-17}

\textbf{Prompt}
& \textbf{Act.} & \textbf{$\Delta$\%} & \textbf{$d$} & \textbf{$t$}
& \textbf{Infl.} & \textbf{$\Delta$\%} & \textbf{$d$} & \textbf{$t$}
& \textbf{Act.} & \textbf{$\Delta$\%} & \textbf{$d$} & \textbf{$t$}
& \textbf{Infl.} & \textbf{$\Delta$\%} & \textbf{$d$} & \textbf{$t$} \\

\midrule

Baseline
& 0.298 & \gradcell{0.0} & --- & ---
& 3.086 & \gradcell{0.0} & --- & ---
& 0.031 & \gradcell{0.0} & --- & ---
& 0.492 & \gradcell{0.0} & --- & --- \\

Polite
& 0.140 & \gradcell{-53.0} & -0.046 & -1.94
& 1.713 & \gradcell{-44.5} & -0.041 & -1.88
& 0.126 & \gradcell{306.5} & 0.093 & 3.94***
& 0.952 & \gradcell{93.5} & 0.071 & 3.28** \\

Random
& 0.160 & \gradcell{-46.3} & -0.039 & -1.79
& 1.656 & \gradcell{-46.3} & -0.036 & -1.73
& 0.146 & \gradcell{371.0} & 0.108 & 4.61***
& 0.968 & \gradcell{96.7} & 0.075 & 3.46*** \\

Low Toxic
& 0.163 & \gradcell{-45.3} & -0.033 & -1.58
& 1.803 & \gradcell{-41.6} & -0.028 & -1.41
& 0.122 & \gradcell{293.5} & 0.082 & 3.51***
& 0.760 & \gradcell{54.5} & 0.046 & 2.21* \\

Med. Toxic
& 0.152 & \gradcell{-49.0} & -0.051 & -2.08*
& 1.340 & \gradcell{-56.6} & -0.047 & -2.01*
& 0.139 & \gradcell{348.4} & 0.097 & 4.08***
& 1.000 & \gradcell{103.3} & 0.058 & 2.72** \\

High Toxic
& 0.147 & \gradcell{-50.7} & -0.056 & -2.24*
& 1.431 & \gradcell{-53.6} & -0.051 & -2.11*
& 0.143 & \gradcell{361.3} & 0.104 & 4.42***
& 0.993 & \gradcell{101.8} & 0.063 & 2.96** \\
\end{tabular}}
\caption{Node-level Activation (Act.) and Influence (Infl.) changes on \gsm{}. $\Delta$\% denotes percentage change relative to the baseline condition (\hlpos{teal}: positive; \hlneg{pink}: negative; shading indicates magnitude), along with Cohen's $d$ and $t$-tests (* $p$<0.05, ** $p$<0.01, *** $p$<0.001).}
\label{tab:node_activation_influence_gsm8k}
\end{table*}

\begin{table*}[!t]
\centering
\footnotesize
\renewcommand*{\arraystretch}{0.9}
\resizebox{\textwidth}{!}{
\begin{tabular}{lcccccccccccccccc}

& \multicolumn{8}{c}{\textbf{Core Nodes}} 
& \multicolumn{8}{c}{\textbf{Variant Nodes}} \\
\cmidrule(r){2-9} 
\cmidrule(l){10-17}

\textbf{Prompt} 
& \textbf{Act.} & \textbf{$\Delta$\%} & \textbf{$d$} & \textbf{$t$}
& \textbf{Infl.} & \textbf{$\Delta$\%} & \textbf{$d$} & \textbf{$t$}
& \textbf{Act.} & \textbf{$\Delta$\%} & \textbf{$d$} & \textbf{$t$}
& \textbf{Infl.} & \textbf{$\Delta$\%} & \textbf{$d$} & \textbf{$t$} \\

\midrule

Baseline 
& 0.264 & \gradcell{0.00} & --- & ---
& 2.914 & \gradcell{0.00} & --- & ---
& 0.047 & \gradcell{0.00} & --- & ---
& 0.973 & \gradcell{0.00} & --- & --- \\

Polite 
& 0.159 & \gradcell{-39.8} & -0.051 & -2.11*
& 1.667 & \gradcell{-42.8} & -0.046 & -2.02*
& 0.102 & \gradcell{+117.0} & 0.074 & 2.98**
& 1.289 & \gradcell{+32.5} & 0.069 & 3.14** \\

Random 
& 0.166 & \gradcell{-37.1} & -0.041 & -1.86
& 1.660 & \gradcell{-43.0} & -0.038 & -1.79
& 0.117 & \gradcell{+148.9} & 0.095 & 3.92***
& 0.788 & \gradcell{-19.0} & -0.026 & -1.31 \\

Low Toxic 
& 0.183 & \gradcell{-30.7} & -0.029 & -1.42
& 1.724 & \gradcell{-40.8} & -0.033 & -1.63
& 0.091 & \gradcell{+93.6} & 0.059 & 2.44*
& 0.931 & \gradcell{-4.3} & -0.012 & -0.61 \\

Med. Toxic 
& 0.158 & \gradcell{-40.2} & -0.056 & -2.29*
& 1.396 & \gradcell{-52.1} & -0.064 & -2.71**
& 0.122 & \gradcell{+159.6} & 0.101 & 4.06***
& 1.258 & \gradcell{+29.3} & 0.044 & 2.17* \\

High Toxic 
& 0.155 & \gradcell{-41.3} & -0.061 & -2.48*
& 1.675 & \gradcell{-42.5} & -0.052 & -2.23*
& 0.119 & \gradcell{+153.2} & 0.091 & 3.71***
& 1.094 & \gradcell{+12.4} & 0.034 & 1.94 \\

\end{tabular}}
\caption{Node-level Activation (Act.) and Influence (Infl.) changes on \mml{}. $\Delta$\% denotes percentage change relative to the baseline condition (\hlpos{teal}: positive; \hlneg{pink}: negative; shading indicates magnitude), along with Cohen's $d$ and $t$-tests (* $p$<0.05, ** $p$<0.01, *** $p$<0.001).}
\label{tab:node_activation_influence_mmlu}
\end{table*}


\subsection{Implementation and Reproducibility Details}
\label{app:implementation}

\para{Models and Compute.}
\autoref{tab:model_details} lists all models used in this work.
Open-weight models were run locally in bfloat16, and GPT-5-nano and Gemini-2.5-Flash were queried through their official APIs with default settings.
We do not train or fine-tune any model.
All local experiments were run on a single node with four NVIDIA RTX PRO 6000 Blackwell (96\,GB) GPUs.

\para{Packages and Settings.}
Inference uses Hugging Face Transformers (v4.57.3) and PyTorch (v2.13.0), with greedy decoding for local models.
Attribution graphs were computed with circuit-tracer (v0.1.0) using \texttt{max\_n\_logits}=3, \texttt{desired\_logit\_prob}=0.95, and \texttt{max\_feature\_nodes}=1024, without node or edge pruning.
Paired $t$-tests and Cohen's $d$ use SciPy (v1.16.2), and mixed-effects regressions use statsmodels (v0.14.5) \texttt{MixedLM} with random intercepts per question (RQ1) or per model (RQ2).
Toxicity scores are Perspective API \texttt{TOXICITY} scores~\cite{perspectiveapi}.


\end{document}